\documentclass[11pt]{article}
\usepackage[round]{natbib}
\usepackage[paper1]{paper_style/kirin-papers}
\usepackage{caption}
\titleformat{\section}{\sffamily\bfseries\color{KirinAccent}\fontsize{15.2pt}{17.5pt}\selectfont}{\thesection}{0.6em}{}
\titleformat{\subsection}{\sffamily\bfseries\color{KirinInk}\fontsize{12.2pt}{14.2pt}\selectfont}{\thesubsection}{0.55em}{}
\titleformat{\paragraph}[runin]{\sffamily\bfseries\color{KirinInk}\normalsize}{}{0pt}{}[\ifnum\spacefactor>1000\else.\fi]

\newtcolorbox{examplebox}[1][]{enhanced,colback=white,colframe=KirinRule,boxrule=0.5pt,arc=1.5mm,
  left=2mm,right=2mm,top=1.1mm,bottom=1.1mm,before skip=2pt,after skip=2pt,fontupper=\sffamily\scriptsize,#1}

\newtcolorbox{floatscopebox}[1]{kirinbase,unbreakable,title=#1,colback=KirinAccentFaint,colframe=KirinRule,
  borderline west={1.6pt}{0pt}{KirinAccent},fontupper=\small}
\newtcolorbox{summarybox}[1]{kirinbase,unbreakable,title=#1,colback=KirinAccentFaint,colbacktitle=KirinAccentPale,
  coltitle=KirinAccentDark,colframe=KirinAccent,borderline west={2.2pt}{0pt}{KirinAccent},fontupper=\small}

\KirinSetShortTitle{Operational Proto-Introspection}
\hypersetup{pdftitle={Operational Proto-Introspection in Looped Language Models: Reading Correctness Before the Answer},pdfauthor={Jan Kirin}}
\begin{document}
\KirinTitleBlock
  {Preprint \textperiodcentered{} Version 4}
  {Operational Proto-Introspection in Looped Language Models}
  {Reading Correctness Before the Answer}
  {Jan Kirin}
  {\href{https://orcid.org/0009-0005-2681-9027}{ORCID 0009-0005-2681-9027}}
  {November 2026}
  {This version keeps the title of versions 1--3 but is a rewrite. It defines the title's term narrowly and replaces the other coined terms with standard ones. It adds answer-free tests of the pre-answer result, run for this version, one of them on a second cohort with question difficulty in the baseline (Section~\ref{sec:fixed}). It leaves out results that the main argument does not need. Appendix~\ref{app:changes} lists the changes.}

\begin{abstractbox}
Can one tell, while a language model reasons, whether its answer will be right? Looped models are a good place to ask: they run the same layers several times per token, leaving one hidden state per loop for a probe to read. I probe Ouro-RLTT, a 2.6B-parameter looped model, for what these states reveal about how its reasoning will end. I call such information operational proto-introspection: operational because it is defined by what a probe can read, and proto because nothing here shows that the model itself uses it.

In part, yes. Reading the first 48 tokens of reasoning, from answers that state no answer yet, a probe on the hidden states reaches an AUROC of 0.758 on a synthetic logic task, against 0.515 for token confidence. Question difficulty is a rival explanation: on 1,164 further questions, token confidence plus simple features of the question predict about as well. It is not only difficulty: between two answers to the same question, the probe scores the correct one higher 74\% of the time. A fine-tuned language model reading the same tokens does about as well. On GSM8K and on a second checkpoint this within-question test is unresolved.

A probe that ranks finished candidate answers is weak at layer~8 on the first loop, much better on the second, and as accurate as the last layers by the third, on four Ouro checkpoints. The direction it reads is largely shared across loops: a probe trained on the last loop reads the first nearly as well as a probe trained there.

The signal helps decide what to do with finished outputs. It improves abstention over surface features and picks correct answers among candidates better than chance, though surface features pick about as well. Once several samples are finished, a baseline that knows their agreement leaves the answer-free probe nothing measurable to add. None of my attempts to use it to make the model generate better answers worked.
\end{abstractbox}

\vspace{2pt}
\noindent\begin{minipage}{\linewidth}
\centering\captionsetup{skip=8pt}
\begin{tcolorbox}[kirinbase,unbreakable,colback=white,colframe=KirinAccent,left=2.2mm,right=2.2mm,top=1.8mm,bottom=1.2mm,before skip=0pt,after skip=2pt]\includegraphics[width=\linewidth]{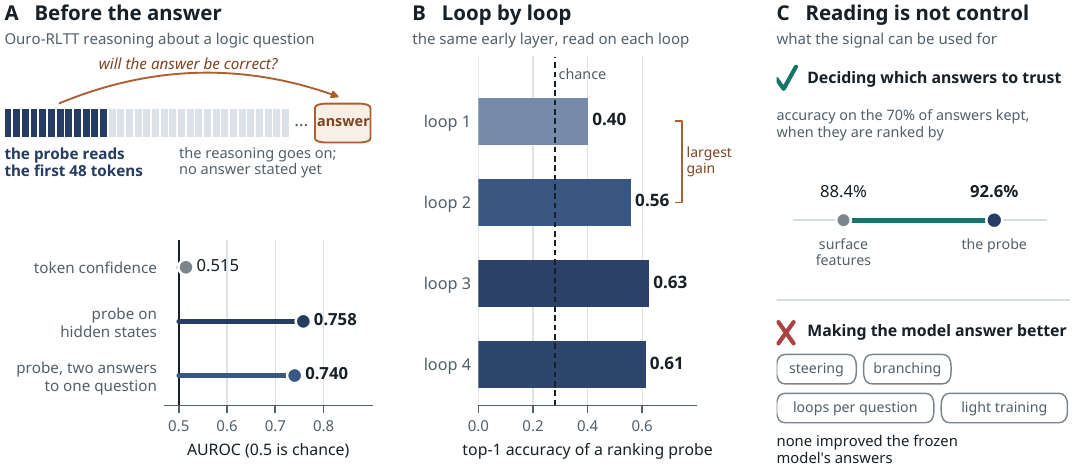}\end{tcolorbox}
\captionof{figure}{\textbf{The three findings.} (A)~A probe reads the hidden states over the first 48 tokens of reasoning, from answers that state no answer yet, and predicts whether the eventual answer will be correct: an AUROC of 0.758, against 0.515 for token confidence. Question difficulty is a rival explanation across questions, but between two answers to the same question the probe still scores the correct one higher (0.740 on a second set of questions; Section~\ref{sec:correct}). (B)~A probe that ranks finished candidate answers is weak at layer~8 on the first loop and as accurate as the last layers by the third (Section~\ref{sec:loops}). (C)~Ranking answers with the probe raises the accuracy of the answers kept, but none of my attempts to make the frozen model generate better answers worked (Section~\ref{sec:use}).}
\label{fig:hero}
\end{minipage}
\clearpage

\section{Introduction}\label{sec:intro}

Watch a student work through a problem, and you can often tell how it will end before they write the answer. They hesitate, or rush, or go back over a step. Does a language model give anything like that away? When it works through a problem, is there any sign, before it answers, of whether the answer will be right?

Looped models are a good place to look. A standard transformer computes each token in a single pass through its layers, like a reader who goes down a page once. A looped model goes down the same page several times. Ouro, for example, runs its 48 layers four times on every token before predicting the next one \citep{zhu2025ouro}. Each pass, which I call a \emph{loop}, refines the hidden state left by the one before, and more loops make the model better at reasoning \citep{saunshi2025looped,geiping2025huginn,zhu2025ouro}.

Because every loop leaves hidden states at every layer, a looped model can be read at many points in its computation of a single token. I read them with probes, small classifiers trained on the hidden states. A probe works like a stethoscope: it listens to the computation from outside, without asking the model anything, and what it hears need not be anything the model itself hears. I ask two practical questions of those states. While the model reasons, do they indicate whether the reasoning is going well? And can that indication be put to use?

For standard transformers the first question has a partial answer. Language models can be trained to predict whether they know the answer to a question \citep{kadavath2022know}, and probes on their hidden states predict whether a statement is true or an answer wrong \citep{azaria2023lying,orgad2024know,kossen2024sep}, whether a chain of thought will succeed before it is written \citep{afzal2025knowing,david2025temporal}, and whether an intermediate answer in a reasoning trace is right \citep{zhang2025reasoning}. A looped model adds a new axis. The same layer can be read on every loop, so I can ask not only whether information is readable, but at which point of the refinement it becomes readable, and whether a probe trained on one loop reads it on the others.

I call such an indication \emph{operational proto-introspection}. It is operational because it is defined by what a probe can extract from the hidden states, not by anything the model reports. It is proto because it is at most a precursor of introspection: it points to how the computation will end, but nothing here shows that the model accesses it (Section~\ref{sec:related}).

I study Ouro-RLTT, Ouro-2.6B-Thinking further trained with reinforcement learning \citep{williams2026rltt}, with its weights frozen. I find three things (Figure~\ref{fig:hero}).

\paragraph{The hidden states predict whether the answer will be correct, before the model commits to one} I take the first 48 tokens of reasoning from answers that state no answer yet. A probe on their hidden states predicts whether the eventual answer is correct with an AUROC of 0.758; the model's token confidence gives 0.515 (Section~\ref{sec:correct}). The test was fixed before its data were generated.

Question difficulty is the obvious rival explanation. A teacher who predicts a wrong answer whenever the problem is hard will often be right without watching the student at all, and a probe could succeed the same way. On a second set of 1,164 questions with deeper proofs, token confidence plus simple facts about the question predicts correctness as well as the hidden states do, and the hidden states add about 0.05 on top, an increase the test leaves unresolved. How much of the probe's signal is difficulty, this comparison cannot say.

The signal is not only difficulty. Take two answers to the same question, one correct and one incorrect, so that difficulty is the same for both: the probe scores the correct one higher 74\% of the time (73 held-out pairs). On the second set this within-question signal tracks timing: it is strongest where the incorrect answer will commit first.

The hidden states are not the only place to find this information. A fine-tuned language model reading the same tokens predicts correctness about as well as the probe across questions on both sets (on the first, under its original labels), and within questions on the second, so much of it is also in the visible text. Nor is the within-question result established elsewhere: on GSM8K, where the model is right 95\% of the time, and on a second checkpoint, Ouro-2.6B-Thinking, the test is unresolved.

\paragraph{Most of the readability arrives with the second loop} For the loop analysis I use a separate task with far more data: ranking finished candidate answers by quality. A probe that ranks them is weak at layer~8 on the first loop (top-1 accuracy 0.40, chance 0.28), much better on the second (0.56), and as accurate as late-layer probes by the third (0.63; Section~\ref{sec:loops}). The rise holds on four Ouro checkpoints, and probes trained on one checkpoint work on its siblings without retraining. In Huginn, a recurrent model of a different design, readability also rises with recurrence.

The direction the probe reads is largely the same on every loop. Probes trained on loop~2 or~3 work on loop~4 without measurable loss. Probes trained on loop~1 lose 0.11--0.21 there, which at first suggests that the first loop stores the information differently. It does not, for the most part: a probe trained on loop~4 reads loop~1 about as well as a probe trained on loop~1 (within 0.02 at layers 16 and~24; 0.05 lower at layer~8). The simplest reading is that the first loop holds the information weakly, like a faint first sketch of the same picture, so a probe fitted there estimates its direction poorly.

\paragraph{The signal helps decide what to do with outputs, but not how to produce them} Letting the probe decide when to abstain raises accuracy on the accepted answers by 4.2 points at 70\% coverage. Letting it rerank finished answers beats random choice, although surface features alone do about as well (Section~\ref{sec:use}). And once several samples of the same question are finished, a baseline that knows their agreement leaves the early probe nothing measurable to add to abstention.

Using the probe inside the loop, to decide which branches of the computation continue, is the most direct use of all. In my search, though, the branches so rarely differed in outcome that the test says nothing either way. Every attempt to improve generation itself failed: steering along probe-derived directions, branching from perturbed states, giving more loops to the questions the probe marks as hard, and training a small adapter to respond to a direction. So far these signals behave like a fuel gauge, which says something about how the trip will end: in my attempts, pushing the needle did not fill the tank.

\medskip
A standard, non-looped transformer also supports probes that rank candidate answers (Section~\ref{sec:nonlooped}), so looping is not what makes this information readable. What looping adds is a view of it developing: the same weights, read four times.

A few caveats apply throughout.
\begin{itemize}
\item I study one model family, at 2.6B parameters and below, with frozen weights. The positive answer-free results of Section~\ref{sec:correct} come from one task family, the logic task, and one checkpoint.
\item The probes read the model from outside. I do not test whether the model itself uses this information, and nothing here shows that it does. In the sense of \citet{lindsey2025introspection}, this is not introspection, which is why I call it proto-introspection.
\item The negative results of Section~\ref{sec:use} hold for the methods, magnitudes and sample sizes I tried. They do not show that no method can work.
\end{itemize}

\begin{figure}[t]
\centering
\begin{minipage}[t]{0.405\linewidth}\vspace{0pt}
\centering
\begin{tikzpicture}[x=1cm,y=1cm,every node/.style={font=\sffamily\scriptsize,text=KirinInk}]
  \node[anchor=west,font=\sffamily\bfseries\footnotesize,text=KirinAccentDark] at (-0.75,4.55) {A\quad Ouro runs its layers four times};
  \foreach \l in {8,16,24,36,47}{
    \node[text=KirinMuted,font=\sffamily\tiny] at ({0.2+4.3*\l/48},3.86) {\l};
  }
  \node[text=KirinMuted,font=\sffamily\tiny,anchor=east] at (0.12,3.86) {layer};
  \foreach \t/\y in {1/3.35,2/2.5,3/1.65,4/0.8}{
    \node[anchor=east,text=KirinMuted] at (0.12,\y) {loop \t};
    \draw[rounded corners=1.2mm,fill=KirinAccentFaint,draw=KirinRule] (0.2,\y-0.2) rectangle (4.5,\y+0.2);
    \foreach \l in {8,16,24,36,47}{
      \fill[KirinAccent] ({0.2+4.3*\l/48},\y) circle (1.7pt);
    }
  }
  \foreach \ya/\yb in {3.35/2.5,2.5/1.65,1.65/0.8}{
    \draw[-{Latex[length=1.6mm]},draw=KirinAccentDark,line width=0.6pt] (4.55,\ya-0.02) .. controls (4.95,\ya-0.1) and (4.95,\yb+0.1) .. (4.55,\yb+0.02);
  }
  \node[anchor=west,text=KirinAccentDark,font=\sffamily\tiny,align=left] at (4.9,2.93) {same\\weights};
  \fill[KirinAccent] (0.3,0.05) circle (1.7pt);
  \node[anchor=west,text=KirinMuted] at (0.4,0.05) {hidden state read by a probe};
\end{tikzpicture}
\end{minipage}\hfill
\begin{minipage}[t]{0.575\linewidth}\vspace{0pt}
{\sffamily\bfseries\footnotesize\color{KirinAccentDark}B\quad What the final-answer-line test reads}\par\smallskip
\begin{examplebox}
\textbf{Question} (Horizon Logic, proof depth 2)\quad Facts: it is not the case that the engine starts; it is not the case that it is raining. Rules: If the engine starts, then it is raining. If it is not the case that the valve is closed, then it is raining. Based only on the facts and rules, is the following statement True, False, or Unknown? Statement: the valve is closed.
\end{examplebox}
\begin{examplebox}[colback=KirinAccentFaint,colframe=KirinAccent!55,borderline west={2pt}{0pt}{KirinAccent}]
\textbf{Answer before the cut: read by the probe}\quad Answer: The given facts are: 1.~It is not the case that the engine starts. 2.~It is not the case that it is raining. The rules are: 1.~If the engine starts, then it is raining. 2.~If the valve is not closed, then it is raining. From the facts, it is not raining. From the second rule, if the valve is not closed, then it is raining. Since it is not raining, the valve must be closed. \textbf{Therefore, the statement ``the valve is closed'' is True.}\par
\textcolor{KirinUnresolved}{\textbf{$\uparrow$ the conclusion is already stated before the cut}}
\end{examplebox}
\begin{examplebox}[fontupper=\sffamily\scriptsize\color{KirinMuted}]
\textbf{After the cut: not read}\quad FINAL ANSWER: A\hfill(correct)
\end{examplebox}
\end{minipage}
\caption{\textbf{What I read.} (A)~Ouro applies the same 48 layers four times per token. Dots mark the layers probed in this paper, which I read on every loop. (B)~A question from the Horizon Logic task and one of four answers sampled from Ouro-RLTT, shown verbatim apart from blank lines; of the other three samples, one was wrong and two gave no parseable answer. The older test of Section~\ref{sec:correct} reads everything before the final-answer line, which here already states the conclusion (Section~\ref{sec:cut}); the answer-free test reads only the first 48 generated tokens of answers that have not committed to an answer (Section~\ref{sec:fixed}).}
\label{fig:overview}
\end{figure}

\section{Setup}\label{sec:setup}

\subsection{Models}
Ouro-RLTT is a 2.6B-parameter looped transformer: 48 layers with hidden size 2,048, applied four times per token, with the output of the last layer on one loop feeding the first layer on the next \citep{zhu2025ouro}. It is Ouro-2.6B-Thinking further trained with RLTT, a reinforcement-learning method that rewards the latent trajectory across loops, at a fixed four loops \citep{williams2026rltt}. Its weights are never changed. The parameters I train are probes, small steering adapters (Section~\ref{sec:steering}) and, in two small experiments, LoRA adapters added to the frozen weights (Section~\ref{sec:training}). I refer to a point in this computation as \emph{layer $\ell$ at loop $t$}, for example layer~8 at loop~3.

For comparison I use three other checkpoints from the same family (Ouro-2.6B, Ouro-2.6B-Thinking, and Ouro-1.4B, which has 24 layers and four loops); Huginn-0125, a 3.5B recurrent model of a different design in which a four-layer block repeats a variable number of times between fixed input and output layers, 32 on average in training \citep{geiping2025huginn}, of which I read the first eight; and MiniCPM-2B-sft, a standard 40-layer transformer \citep{hu2024minicpm}.

\subsection{Features and probes}
I read the residual stream after a given layer on a given loop and average it over token positions. Layers 8, 16, 24 and~36 are the outputs of those blocks; layer~47 is the state at the end of each loop, after Ouro's final normalization, which is what the next loop receives. Unless stated otherwise I concatenate layers 24, 36 and~47 on all four loops, giving a $3\times4\times2{,}048=24{,}576$-dimensional vector; these three layers were chosen in earlier work on this model. Section~\ref{sec:loops} adds layers 8 and~16. I use two kinds of probe, both linear.

A \emph{correctness probe} maps the features of one answer to the probability that the answer is correct. Features are standardized and reduced with PCA before an L2-regularized logistic regression, and the number of components and the regularization strength are chosen by cross-validation grouped by question, on training and validation questions only. Section~\ref{sec:correct} trains it on the text before the final-answer line and, as the answer-free probe, on the first tokens of reasoning; Section~\ref{sec:use} also trains one on complete answers.

A \emph{ranking probe} compares two candidate answers $A$ and $B$ to the same question with the score $s(A,B)=w^\top(h_A-h_B)$, where $h$ are the features and there is no bias term. Swapping the candidates flips the sign of the score, so the probe cannot prefer a candidate for the position in which it was presented. Section~\ref{sec:loops} and the pruning of Section~\ref{sec:prune} use it. The steering directions of Section~\ref{sec:steering} come from probes of earlier work on this model.

\subsection{Tasks}
\paragraph{GSM8K} 170 grade-school math problems \citep{cobbe2021gsm8k} with four sampled solutions each (680 answers, 407 correct), labeled by comparing the final number with the reference.

\paragraph{Horizon Logic} Synthetic propositional-logic questions (Figure~\ref{fig:overview}B). Each gives a few facts and rules and asks whether a statement is True, False or Unknown. The proof depth, the number of rule applications needed, is set between 2 and~4, and a truth-table checker provides the label. The prompt asks the model to end with the line ``FINAL ANSWER: $\langle$letter$\rangle$''. I sampled four answers per question (temperature 0.7, top-$p$ 0.95, top-$k$ 50, at most 448 new tokens) for 170 questions, and later for 510 more generated the same way (Section~\ref{sec:results3}). Token confidence is the log-probability of each sampled token under this sampling distribution.

\paragraph{Candidate ranking} Groups of candidate answers in five domains (alignment, reasoning, math, coding and logic), labeled for quality and drawn from public datasets. Some wrong candidates are real mistakes; in coding and math many are constructed by corrupting a correct answer. Section~\ref{sec:loops} uses a fixed subset of 250 training, 60 validation and 120 held-out groups per domain (8,592 candidates).

\subsection{Evaluation}\label{sec:eval}
All splits are by question: no question contributes answers to both training and test data, and this is checked in code. Confidence intervals are 95\% intervals from a bootstrap that resamples questions rather than answers, because answers to the same question are not independent; paired comparisons reuse the same resamples for both arms. When a probe picks one answer from a group, I compare it with random choice, whose chance of success I compute exactly from the number of correct answers in each group. If several answers tie for the probe's top score, the pick earns the share of them that are best.

Two errors inflated results in an earlier paper on this model \citep{kirin2026rpe} and in earlier versions of this one. The first was group leakage \citep{kapoor2023leakage}: several training rows were built from each source item, the rows were split at random, and siblings of test rows ended up in training. The second was position bias \citep{zheng2023judge,wang2023fair}: a pairwise scorer trained with the candidates in random order but tested with the preferred candidate always shown first learned to prefer the first candidate. Splitting by question and scoring both orders removes both. Appendix~\ref{app:changes} notes what this changed. The box ``Terms used throughout'' defines the terms used from here on.

\begin{table}[!tb]
\begin{floatscopebox}{Terms used throughout}
\textbf{AUROC.} The chance that a probe scores a randomly chosen correct answer above a randomly chosen incorrect one: 0.5 is chance, 1 is perfect.\par
\textbf{Token confidence.} The log-probability of the tokens the model sampled, under the sampling distribution.\par
\textbf{Surface features.} What can be measured without reading hidden states, such as an answer's length and its token confidence.\par
\textbf{Answer-free.} Text in which the detector of Section~\ref{sec:fixed} finds no stated answer. An answer \emph{commits} where it first gives its final answer in one of the labeled formats.\par
\textbf{Across and within questions.} Across questions, the answers to all questions are compared together, so a probe can succeed by recognizing hard questions. Within a question, a correct and an incorrect answer to the same question are compared, so it cannot.\par
\textbf{Held out, out of fold.} Held-out questions are never used in training. Out-of-fold scoring trains several probes, each without one share of the questions, and scores every answer with a probe that was not trained on its question.\par
\textbf{Unresolved.} The 95\% interval is too wide to show an effect or to rule one out.
\end{floatscopebox}
\end{table}

\section{Hidden states predict whether the answer will be correct}\label{sec:correct}

Do the hidden states reveal whether the model's answer will be correct, before it gives that answer? Two cheap signals already predict this. Long solutions go wrong more often, and so do solutions the model writes with low confidence. The question is whether the hidden states add anything to these. I test it twice: at the final-answer line, as earlier versions of this paper did, and early in the reasoning, before the model has committed to any answer. A third cheap signal, question difficulty, turns out to matter early in the reasoning: with it in the baseline the hidden states add little across questions, yet they still separate two answers to the same question.

\subsection{At the final-answer line}\label{sec:results3}
For each answer I cut the text at the first occurrence of the final-answer marker, run the frozen model on the prompt plus the text before the cut, and read the hidden states (Figure~\ref{fig:overview}B). On GSM8K the cut came earlier when the reference number appeared first, which Section~\ref{sec:cut} shows is a problem. As baselines I use surface features of the same text: on GSM8K three length features and three confidence features (Appendix~\ref{app:correct}), and on Horizon Logic its length, mean and minimum token log-probability and whether the answer hit the token limit.

On GSM8K, the length features give an AUROC of 0.687 and the confidence features 0.569; together they give 0.731 (Table~\ref{tab:correct}). Hidden states alone give 0.745, and adding them to the surface features gives 0.797, an increase of 0.066 (95\% CI $[0.021, 0.112]$), measured on out-of-fold predictions grouped by question.

On Horizon Logic the same test gives a larger increase, on three sets of questions. About a quarter of the sampled answers never produce a parseable answer, usually because they run out of tokens on deeper proofs, and I leave them out of the labels. On the first 170 questions the hidden states added 0.141, but with only about 19 incorrect held-out answers the interval was wide ($[0.004, 0.290]$). On 510 new questions generated the same way the hidden states added 0.095 ($[0.038, 0.157]$). Pooled over all 680 questions the increase is 0.111 ($[0.056, 0.169]$). On a third, fresh set of 599 questions generated for this version (Section~\ref{sec:fixed}), the same protocol gives 0.095 again ($[0.012, 0.177]$). A check run after these results adds the question's proof depth and prompt length to the baseline; the increase then holds on all 680 questions ($+0.106$, $[0.054, 0.165]$) but not on the 599 fresh ones ($+0.032$, $[-0.031, 0.098]$).

\begin{table}[t]
\centering
\caption{Predicting correctness from hidden states (Ouro-RLTT unless noted). AUROC of a probe on surface features, on hidden states, and on both, with the increase from adding hidden states. At 48 tokens the surface features are the mean and minimum log-probability of those tokens; the question features of the second set are prompt length, proof depth, the numbers of variables, facts and rules, and the correct option; the problem features on GSM8K are prompt length and the number of steps in the reference solution. Where Both equals Hidden, the combined probe compressed hidden and surface features together (Section~\ref{sec:fixed}). In the end-of-prompt row, the Surface column is the baseline with that state added.}
\label{tab:correct}
\small
\begin{tabular}{@{}lrrrl@{}}
\toprule
 & Surface & Hidden & Both & Increase [95\% CI] \\
\midrule
\multicolumn{5}{@{}l}{\emph{At the final-answer line}} \\
GSM8K (170 questions) & 0.731 & 0.745 & 0.797 & $+0.066$ $[+0.021, +0.112]$ \\
Horizon Logic, all 680 questions & 0.652 & 0.763 & 0.763 & $+0.111$ $[+0.056, +0.169]$ \\
Horizon Logic, 599 fresh questions & 0.687 & 0.782 & 0.782 & $+0.095$ $[+0.012, +0.177]$ \\
\addlinespace
\multicolumn{5}{@{}l}{\emph{48 tokens into the reasoning, before any answer (599 fresh questions)}} \\
Answers stating no answer yet & 0.515 & 0.758 & 0.758 & $+0.243$ $[+0.129, +0.365]$ \\
Same, labeled by the first final answer (after the result) & 0.523 & 0.777 & 0.777 & $+0.254$ $[+0.105, +0.407]$ \\
\addlinespace
\multicolumn{5}{@{}l}{\emph{48 tokens, second set (1,164 questions at depth 3--4, labeled by the first final answer)}} \\
Log-probabilities only, as above & 0.530 & 0.683 & 0.683 & $+0.153$ $[+0.047, +0.259]$ \\
Log-probabilities and question features & 0.681 & 0.683 & 0.734 & $+0.053$ $[-0.006, +0.109]$ \\
Same, end-of-prompt state in the baseline & 0.709 & 0.683 & 0.758 & $+0.049$ $[+0.008, +0.089]$ \\
\addlinespace
\multicolumn{5}{@{}l}{\emph{32 tokens, second set, Ouro-2.6B-Thinking (same questions)}} \\
Log-probabilities and question features & 0.689 & 0.653 & 0.709 & $+0.021$ $[-0.034, +0.076]$ \\
\addlinespace
\multicolumn{5}{@{}l}{\emph{96 tokens, GSM8K training split (2,000 problems, labeled by the first final answer)}} \\
Log-probabilities and problem features & 0.620 & 0.632 & 0.678 & $+0.059$ $[-0.039, +0.156]$ \\
\bottomrule
\end{tabular}
\end{table}

\subsection{Why the final-answer line is not enough}\label{sec:cut}
The final-answer line is not the first place an answer appears. Models often state their conclusion in the reasoning and then repeat it on the answer line, as in Figure~\ref{fig:overview}B. An audit run after version~3 of this paper counted how often: of the 1,557 parseable answers to the 510 new Horizon questions, 871 (56\%) state the correct conclusion before the cut, and 89.7\% of those questions (451 of 503) have at least one such answer.

The cut can also fall too early: the model often writes an empty ``FINAL ANSWER:'' template before it starts reasoning, and the cut then removes the reasoning with it. So the test above cannot separate a forecast of success from a check of an answer already given, and it cannot support the description earlier versions used, ``strictly pre-answer''. Testing that needs a cut placed before any answer.

GSM8K has a second problem. Its cut was placed at the final-answer marker or at the first appearance of the reference number, whichever came first, so that the number could not leak into the text read. But the label is computed from the same number, so where the text ends depends on the label being predicted. Rerunning the protocol on 600 other test problems, with the generated tokens kept, exposed a third problem: many labels reflect the token budget, not correctness. With the original limit of 224 new tokens, 60.6\% of answers count as correct; with 448 tokens, 94.2\% of the model's parseable answers are correct.

\subsection{Before the model commits to an answer}\label{sec:fixed}
I therefore generated a fresh set of 600 Horizon Logic questions (599 completed) with the same generator, prompts and sampling, this time keeping the exact generated tokens. For each answer I take the hidden states over the prompt and the first $N$ generated tokens, for $N$ from 16 to 128.

The rest of this section follows one route. A first set of questions shows a large gain. I then test two simpler explanations, the visible text and question difficulty. A second set of questions, built to test difficulty, shows that a baseline with question features predicts about as well as the hidden states across questions, and that a signal still holds between two answers to the same question.

A detector flags answers that already state an answer within those tokens: an option after the final-answer marker, ``the answer is C'', ``the statement is True'', ``cannot be determined'', a boxed answer, a line that is only an option, or the question's own statement asserted after words such as ``therefore'' or ``final answer''. Restated option lists (``True, False, or Unknown'') and the empty answer template do not count.

I fixed the detector and every threshold before generating any data, and the analysis code before opening any of it. The cut was chosen without labels, as the longest prefix at which at least 75\% of training answers state no answer: 48 tokens. Before opening any label I read 20 of these training prefixes; none stated an answer. At a fixed cut every answer has the same length, so the surface features are the mean and minimum log-probability of the tokens read. The test covers answers that state no answer within those tokens and later give an answer the parser accepts.

\begin{figure}[t]
\centering
\includegraphics[width=\linewidth]{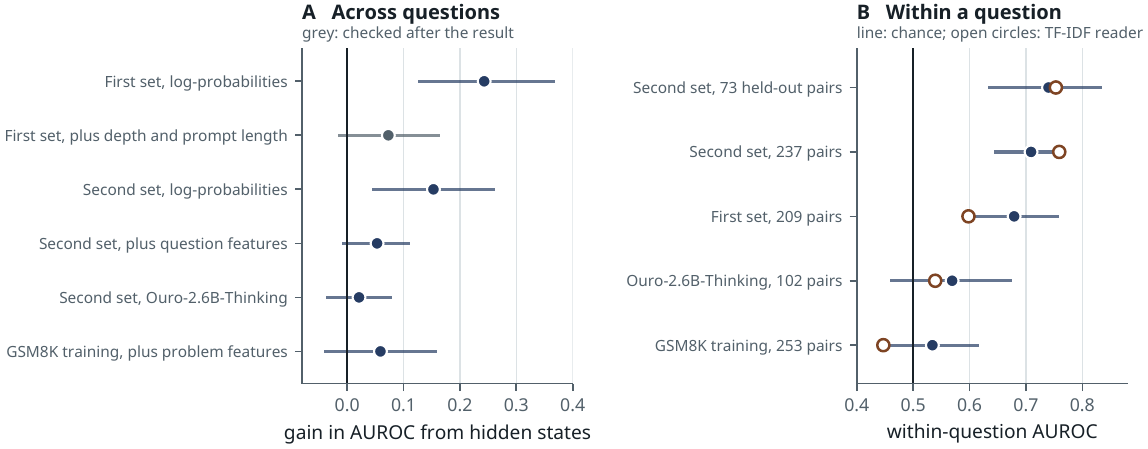}
\caption{\textbf{Predicting correctness before the answer} (answers that state no answer by the cut; 95\% question-clustered intervals). (A)~Across questions: the gain in AUROC from adding hidden states to the baseline. Against the log-probabilities alone the gain is large; with question or problem features in the baseline it is small and unresolved. The first set's difficulty check was run after its result. (B)~Within questions: how often the probe scores the correct answer of a mixed pair above the incorrect one, every answer scored out of fold (the second set also on its held-out pairs). Open circles: a TF-IDF reader of the same tokens, scored the same way. In~(B), only the second set's held-out pairs and the Ouro-2.6B-Thinking test were fixed in advance.}
\label{fig:correct}
\end{figure}

\paragraph{The first set: a large gain} Among answers that state no answer within 48 tokens (412 of the 530 parseable held-out answers, from 165 questions, 45 of them incorrect), a probe on the hidden states predicts whether the eventual answer is correct with an AUROC of 0.758. The surface features give 0.515, so the hidden states add 0.243 ($[0.129, 0.365]$; Figure~\ref{fig:correct}A).

The gain appears at every cut from 16 to 96 tokens. It holds when only questions whose four answers all state nothing yet are kept ($+0.199$, $[0.039, 0.360]$), and when each answer is labeled by its first final answer instead of its last ($+0.254$, $[0.105, 0.407]$). The detector misses one format, a Markdown-bold ``**Final Answer**'' followed by the option, and 15 of these answers state an answer that way inside the window. Without them the increase is $+0.253$ ($[0.135, 0.365]$).

Two simpler explanations come to mind: the visible text, and the difficulty of the question.

\paragraph{Is it the visible text?} A TF-IDF classifier on the prompt and the same 48 tokens reaches an AUROC of 0.641, and the hidden states beat it by 0.118 ($[0.032, 0.206]$). A fine-tuned language model is a stronger reader. After the results I tried two, with every choice fixed in advance (Appendix~\ref{app:correct}). A fine-tuned DeBERTa-v3-large reaches 0.770, level with the probe: the probe's lead is $-0.012$ ($[-0.081, 0.054]$), so no difference is resolved. Relabeled by the first final answer, the same reader reaches only 0.523 (Appendix~\ref{app:correct}). A Qwen2.5-1.5B tuned with LoRA reaches 0.552, and the probe beats it by 0.206 ($[0.074, 0.324]$). Across questions, then, the hidden states do not beat every reader of the visible text.

\paragraph{Is it difficulty?} Deep proofs and long prompts go wrong more often, and the surface features above include neither. Measuring this needs a fix to the combined probe. Earlier versions compressed hidden states and surface features together with PCA. With 24,576 hidden dimensions against a handful of surface features, the surface features kept almost no weight (0.011\% in the first 170 questions), and the combined probe scored like the hidden-only one. From here on I compress only the hidden states and add the surface features afterwards.

With this fix, proof depth and prompt length alone reach an AUROC of 0.686 on these answers, against the hidden states' 0.758, and adding them to the log-probabilities leaves an increase of 0.073 ($[-0.014, 0.162]$). I ran these checks after seeing the result, so I tested difficulty again on new data, with the analysis fixed in advance.

\paragraph{The second set} It has 1,164 new questions, all at proof depth 3 or~4, the two deepest levels of the task family. Generation was the same as before. Two things changed, both fixed before any data existed.

First, each answer is labeled by its first final answer: the first option letter after a final-answer marker or in a box, or failing that the first option word after a marker. On these data, among these formats, this is always the earliest answer the text gives. The model often keeps writing after it answers, and the old parser used the last one.

Second, the baseline adds six facts about the question to the log-probabilities: prompt length, proof depth, the numbers of variables, facts and rules, and which option is correct. The last one is needed because the hidden states at 48 tokens contain which option is correct (Appendix~\ref{app:correct}); it is a fair control, but not something available for a new question. The cut was again 48 tokens, chosen without labels, and a hand audit of 20 prefixes found no stated answer.

Analyzed the old way, the second set reproduces the first: the hidden states add 0.153 ($[0.047, 0.259]$) to the log-probabilities. With the six question features in the baseline, the picture changes. The log-probabilities and question features together reach an AUROC of 0.681, and the hidden states alone 0.683. All of them together reach 0.734, an increase of 0.053 ($[-0.006, 0.109]$; 735 held-out answers from 326 questions, 59 incorrect).

The plan fixed in advance calls this result unresolved; its headline rule, which has only two labels, then reads ``mostly difficulty'' rather than ``beyond difficulty''. The result also sits on a knife edge. The sealed code does not fix how questions are split into cross-validation folds; the split depends on how the machine breaks ties. Over 31 random tie orders the increase ranges from 0.052 to 0.064, with an interval clear of zero in 16 of them (Appendix~\ref{app:correct}). Across questions, then, the hidden states add about 0.05 to a baseline that holds the question features. The smaller increase shows that features of the question predict much of what the probe predicts; it does not measure what share of the hidden states' information is difficulty, and I make no claim about that share. Adding the model's hidden state at the last prompt token to the baseline leaves the first 48 tokens about the same increase ($+0.049$, $[0.008, 0.089]$).

\paragraph{Within a question, a signal remains} The cleanest test compares two answers to the same question, one correct and one incorrect. The question and its difficulty are then identical. The probe scores the correct answer higher in 74\% of the 73 held-out pairs (AUROC 0.740, $[0.636, 0.831]$; 43 questions); the log-probabilities and question features manage 52\%, and a probe on hidden states and baseline together 69\% ($[0.577, 0.791]$).

This result does not depend on the folds (73--74\% under all 31 tie orders). It holds over all 237 such pairs when every answer is scored out of fold (71\%, $[0.646, 0.767]$), and at every cut from 16 to 96 tokens (63--74\%). The first set shows the same: 68\% over 209 pairs ($[0.602, 0.755]$), or 66\% ($[0.568, 0.736]$) without its 15 missed answers (Figure~\ref{fig:correct}B). The checks in this paragraph were run after the result.

\paragraph{The within-question signal tracks timing} On the second set, in 78\% of the 237 pairs, the correct answer states its final answer later than the incorrect one. The probe's within-question score is concentrated in these pairs: 76\% on the 185 where the incorrect answer commits first, against 52\% on the other 52. The commit itself comes after the 48 tokens the probe reads (in all but 9 answers; Appendix~\ref{app:correct}); what the probe picks up in those tokens goes with it.

A TF-IDF reader of the same 48 tokens does as well within questions (75\% on the 73 held-out pairs; 76\% against the probe's 71\% over all 237), and across questions the hidden states gain little over the reader once difficulty is in both baselines ($+0.014$, $[-0.023, 0.050]$). The fine-tuned readers point the same way. On the second set DeBERTa matches the probe within questions (0.719 against 0.740 on the held-out pairs; 0.738 each over all pairs out of fold), while the probe beats the Qwen reader ($+0.095$, $[0.019, 0.170]$, out of fold). On the first set the probe beats both readers within questions out of fold ($+0.096$, $[0.029, 0.163]$, against DeBERTa), a descriptive comparison (Appendix~\ref{app:correct}), but on its 57 held-out pairs no difference from DeBERTa is resolved.

On the second set, then, the probe and the text reader mostly pick out answers that will commit early and turn out wrong. That is consistent with the model hurrying toward a wrong answer. These timing checks were run after the result.

\paragraph{GSM8K} The first GSM8K test had too few errors. With 448 tokens, 94\% of the model's parseable answers to the 600 problems of Section~\ref{sec:cut} are correct, which leaves 21 incorrect held-out answers at the cut, below the minimum of 30 fixed in advance (Appendix~\ref{app:correct}). I therefore ran a second GSM8K test on 2,000 problems from the training split, with the analysis of the second set fixed in advance. Across questions the result resembles the logic task's: the hidden states add 0.059 to token confidence and the problem's features ($[-0.039, 0.156]$; 1,907 held-out answers, 85 incorrect), which the plan calls unresolved.

Within questions the test is unresolved. Scoring every answer out of fold, as for the second set, gives 93 problems and 253 pairs, and the probe scores the correct answer higher in 53\% of them ($[0.450, 0.613]$; run after the result), an interval that includes chance. Commit timing hardly separates GSM8K pairs either (AUROC 0.53, against 0.78 on the logic task).

\begin{summarybox}{Summary: before the model commits}
\begin{itemize}[leftmargin=1.2em,itemsep=1.5pt]
\item Before the model states an answer, its hidden states predict whether that answer will be right.
\item Question difficulty matters: a test without difficulty in its baseline, like the first one here, gives about three times the increase of one with it, although the estimator changed between the two as well.
\item On the logic task the signal also holds with the question fixed, so it is not only difficulty. On the second set it tracks how soon an answer will commit, a check run after the result.
\item A fine-tuned reader of the visible text predicts about as well as the probe across questions on both sets (on the first, under its original labels) and within questions on the second, so much of what the probe reads is also in the text.
\item In Ouro, on the logic task, the prompt alone predicts little beyond difficulty (Appendix~\ref{app:correct}); the first tokens of reasoning predict more.
\end{itemize}
\end{summarybox}

\subsection{A second checkpoint}\label{sec:thinking}
Does the result hold on another checkpoint? On Ouro-2.6B-Thinking the final-answer-line test is inconclusive: 46\% of answers gave no parseable answer within the 448-token limit (against 25\% for Ouro-RLTT), and a follow-up on 900 new questions gave $+0.042$ ($[-0.017, 0.104]$), which neither replicates the Ouro-RLTT result nor rules out an effect of that size.

I then repeated the second set's answer-free test on this checkpoint, with the same 1,164 questions, prompts and sampling seeds, and an analysis fixed in advance whose primary endpoint is the within-question test with every answer scored out of fold. Only 47\% of Ouro-2.6B-Thinking's answers state a final answer, and it is right on 89\% of those, so only 74 questions have both a correct and an incorrect answer. The cut chosen without labels is 32 tokens, because at 48 tokens fewer than 75\% of training answers are still answer-free. Reading 20 of the answers at this cut, most are still extending the prompt's list of options with invented ones rather than reasoning.

Within questions the probe scores the correct answer higher in 57\% of the 102 pairs ($[0.462, 0.673]$), which the plan calls unresolved. Across questions the hidden states add $+0.021$ ($[-0.034, 0.076]$) to the log-probabilities and question features. On the same questions, scored the same way, Ouro-RLTT gives 0.717 ($[0.654, 0.779]$; 237 pairs from 130 questions), 0.149 more than Ouro-2.6B-Thinking ($[0.023, 0.269]$; Appendix~\ref{app:correct}). The two checkpoints' mixed questions differ, and this one's prefixes hold little reasoning; the test cannot separate these from a real difference between the checkpoints, and an unresolved result does not show that the signal is absent.

\section{Readability rises across loops}\label{sec:loops}

Ouro applies the same 48 layers four times. That allows a question with no analogue in a standard transformer: is layer~8 on the fourth loop doing the same thing as layer~8 on the first? If the loops refine one shared representation, a probe trained on one loop should work on the others, in both directions. If they store the information in different forms, it should not.

I answer this with the ranking probe on the candidate-ranking set (Section~\ref{sec:setup}), which has far more data than the reasoning tasks of Section~\ref{sec:correct}. The target therefore differs from Section~\ref{sec:correct}: the probe ranks finished candidate answers by quality, rather than predicting from the model's own early reasoning whether it will be right. At each of layers 8, 16 and~24 on each of the four loops I train a probe with the same grid of 36 hyperparameter settings, choosing among them on validation data only. I then make two measurements. First, I test each probe where it was trained: is the information readable there? Second, I apply each probe, unchanged, to the other loops of the same layer: is the information stored the same way there? I first moved probes only to later loops; I added the moves to earlier loops after those results.

\begin{figure}[t]
\centering
\includegraphics[width=\linewidth]{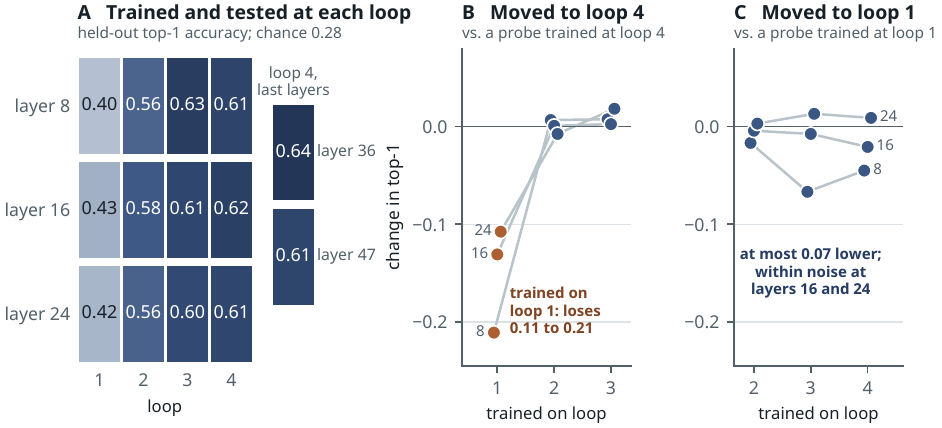}
\caption{\textbf{Readability rises across loops} (Ouro-RLTT). (A)~Held-out top-1 accuracy of a ranking probe trained and tested at each layer and loop. Early layers are weak on loop~1, much better on loop~2, and as accurate as the last layers by loop~3. (B)~The change in accuracy when a probe trained on one loop is applied, unchanged, at loop~4, relative to a probe trained at loop~4. Probes trained on loop~1 lose 0.11--0.21; probes trained on loops 2 or~3 lose nothing measurable. (C)~The reverse move: a probe trained on a later loop applied at loop~1, relative to a probe trained at loop~1. The loss is at most 0.07, and indistinguishable from zero at layers 16 and~24, so the loop-1 probe's poor transfer in~(B) is most simply read as a weak loop-1 signal rather than a different representation. Panel~C was added after the other results. Intervals and the full matrices are in Appendix~\ref{app:loops}.}
\label{fig:loops}
\end{figure}

\subsection{Most of the gain comes with the second loop}
Figure~\ref{fig:loops}A shows held-out top-1 accuracy: the fraction of groups in which the probe's highest-ranked candidate is among the best (chance 0.28). On the first loop every layer is weak: 0.40 at layer~8, 0.43 at layer~16 and 0.42 at layer~24. The second loop brings the largest gain, to 0.56--0.58. By loop~3, layers 8 and~16 reach 0.63 and 0.61, as accurate as probes on the last layers at loop~4 (0.64 at layer~36 and 0.61 at layer~47). From loop~1 to loop~4 accuracy rises by 0.19--0.21 at every layer, with intervals well clear of zero.

\subsection{The direction is shared across loops}
Figure~\ref{fig:loops}B shows what happens when a probe is moved to loop~4. A probe trained on loop~1 loses 0.11--0.21 there, compared with a probe trained at loop~4, and its scores agree only moderately with that probe's (Spearman correlation 0.53--0.78). Probes trained on loop~2 or~3 lose nothing measurable: the change is within $\pm0.02$, and their scores correlate at 0.90--0.99 with those of the loop-4 probe.

On its own, this suggests that loop~1 stores the ranking direction in a different form. Moving probes the other way mostly rules that out (Figure~\ref{fig:loops}C). A probe trained on loop~4 and applied at loop~1 does about as well there as a probe trained at loop~1: the difference is $-0.021$ ($[-0.057, 0.015]$) at layer~16 and $+0.009$ ($[-0.028, 0.045]$) at layer~24. A probe pointing in an unrelated direction would fall to chance, 0.12--0.15 lower. Only at layer~8 do probes from loops 3 and~4 lose a little at loop~1 ($-0.067$, $[-0.108, -0.025]$, and $-0.045$, $[-0.085, -0.003]$). Moves among loops 2, 3 and~4 cost nothing measurable in either direction (Appendix~\ref{app:loops}).

So the direction that ranks candidates is largely shared across all four loops. The simplest explanation is that the first loop holds it weakly, and a probe fitted there is a noisy estimate of it, which is why that probe transfers badly. This is an interpretation: shifts in scale or covariance between loops, components specific to one loop, or how well the probes' short training (60 optimizer steps) fits each loop could produce the same asymmetry. At layer~8 the first loop may also store part of it differently.

\subsection{Checks}
Three checks address simple explanations.
\begin{itemize}
\item \emph{Shuffled labels.} Probes trained on shuffled labels score 0.33--0.42, above the chance rate of 0.28, which already accounts for group sizes and ties. I have not found why. Each control is a single shuffled fit, not a null distribution, so I compare each probe with its own control. Layer~8 at loop~1 beats its control by only 0.06, layer~8 at loop~4 by 0.19: first-loop readability at layer~8 is weak in absolute terms.
\item \emph{Length.} Ranking candidates by length gives 0.26 with the longest first and 0.23 with the shortest first, both below chance.
\item \emph{Extraction.} Re-extracting the layer-24, 36 and~47 features along the new path reproduces the stored ones: a reference probe scores 0.687 on the stored features and 0.685 on the new ones.
\end{itemize}

\subsection{Beyond Ouro-RLTT}
\begin{figure}[t]
\centering
\includegraphics[width=\linewidth]{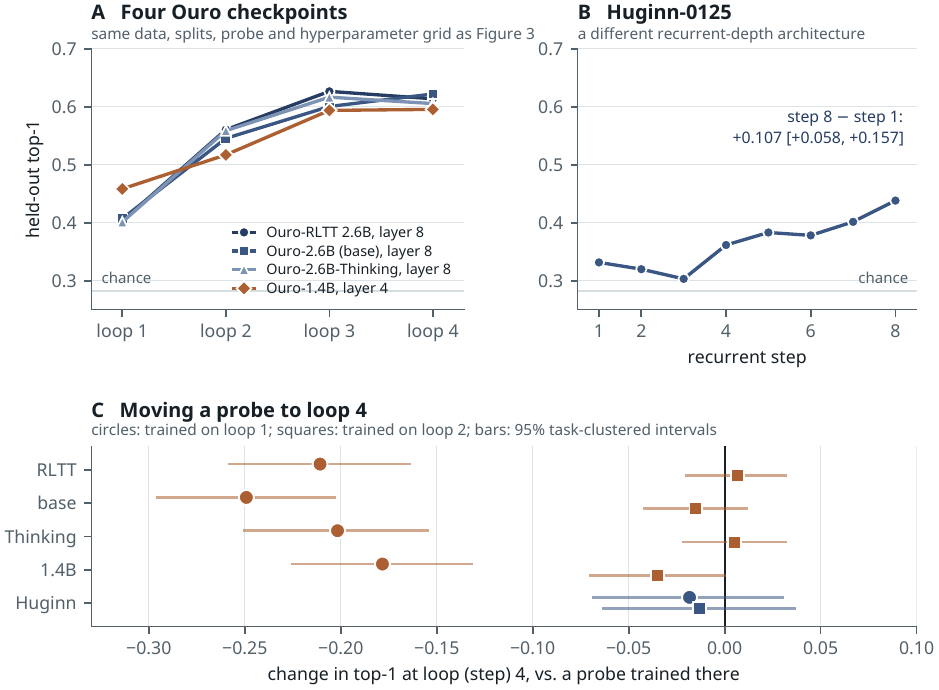}
\caption{\textbf{The pattern beyond Ouro-RLTT.} (A)~Accuracy at the earliest probed layer across loops, for four Ouro checkpoints (layer~4 for the 24-layer Ouro-1.4B). (B)~Huginn-0125 across eight recurrent steps. (C)~The change at loop (or step)~4 for a probe trained on loop~1 (circles) or loop~2 (squares), relative to a probe trained at loop~4, with 95\% intervals. On every Ouro checkpoint the loop-1 probe transfers poorly. In Huginn the step-1 probe loses nothing measurable at step~4, where readability has barely risen; by step~8 it loses 0.093 (Appendix~\ref{app:loops}).}
\label{fig:replication}
\end{figure}

Figure~\ref{fig:replication} repeats the analysis, with the same data and procedure, on other models. On the base and Thinking versions of Ouro-2.6B and on Ouro-1.4B the pattern is the same. Early layers are weak on loop~1 (0.40--0.47) and reach 0.59--0.64 by loops 3 and~4, and at the earliest layer a probe trained on loop~1 loses 0.18--0.25 at loop~4 while a probe trained on loop~2 loses little or nothing (Figure~\ref{fig:replication}C). I did not run the reverse moves on these checkpoints. Probes trained on Ouro-RLTT also work on the base and Thinking checkpoints without retraining, landing within 0.028 of probes trained on those checkpoints, with Spearman correlations between the two probes' scores of 0.87--1.00 (medians 0.97 and 1.00). The structure is shared across the family rather than created by RL training.

One difference comes with scale: on Ouro-1.4B the early layers end above the model's own last layers (0.62 at layer~12 on loop~3, against 0.56 and 0.55 at layers 18 and~23 on loop~4), so ``the last layers read best'' is a property of the 2.6B models, not of looping.

Huginn has a different design, a four-layer block repeated between fixed input and output layers \citep{geiping2025huginn}. Reading the output of the repeated block over eight steps, readability also rises with recurrence, from 0.33 at step~1 to 0.44 at step~8 ($+0.107$, $[0.058, 0.157]$), though at a lower level than in Ouro. Probes trained on step~1 transfer to steps 2--4 without measurable loss, but readability has barely risen at those steps (0.30--0.36). At step~8, where it has, the step-1 probe is 0.093 below a probe trained there ($[-0.142, -0.045]$; Appendix~\ref{app:loops}), which resembles Ouro's loop-1 probes. Eight steps cover only the start of the 32 Huginn uses on average in training. I conclude that the rise with recurrence generalizes, and that these data do not show Huginn's first-step probes transferring better than Ouro's.

\subsection{What the Ouro paper already shows}
Two observations in \citet{zhu2025ouro} fit the large gain from the second loop. When Ouro-1.4B decodes with a single KV cache instead of one per loop, keeping only the last loop's cache, or the average over loops, costs almost nothing, but keeping only the first loop's cache collapses GSM8K accuracy from 78.9 to 18.7 (their Table~14); they did not test the caches of loops 2 and~3 alone. And the largest gain from an extra loop comes with the second: on OlympiadBench, Ouro-2.6B-Thinking scores 19.0 with one loop and 68.6 with two (their Table~13). Their own probes, on a paraphrase task, also find that a new loop can revise the answer suggested by the one before (their Section~7.2). None of these tests my probes directly.

\subsection{Looping is not what makes the signal readable}\label{sec:nonlooped}
Is a looped model needed for any of this? I trained ranking probes of the same kind on MiniCPM-2B-sft, a standard 40-layer transformer, reading layers 24 and~36. This comparison uses the full candidate set, split by task, with 3,525 held-out groups, not the subset used above. MiniCPM reaches 0.544 held-out top-1, against 0.627 for Ouro-RLTT on the same split; I report no chance rate for this split, so the two values compare with each other, not with the 0.28 above. A standard transformer therefore supports this kind of probe as well. The two models differ in training data, objective, width and much else, so I do not attribute the gap to looping. What looping adds is not readability but a view of it: the same layer, read on four loops, shows where and when the information appears.

\section{Using the signal}\label{sec:use}

A signal is only as useful as what can be done with it. Some uses act on outputs the model has already produced: deciding whether to answer at all, and which of several finished answers to return. One acts during the computation, deciding which branches of it continue. The rest try to change what the model produces: steering its hidden states, branching its computation, giving more loops to questions that need them, and training it lightly to respond to a direction. The first kind worked, the second could not be tested properly on my data, and the third did not work. The box ``Seven uses and what came of them'' lists all seven.

\begin{table}[!tb]
\begin{floatscopebox}{Seven uses and what came of them}
\renewcommand{\arraystretch}{1.06}
\begin{tabularx}{\linewidth}{@{}>{\raggedright\arraybackslash}p{0.19\linewidth}>{\raggedright\arraybackslash}p{0.29\linewidth}>{\raggedright\arraybackslash}X@{}}
\textbf{Use} & \textbf{What is tried} & \textbf{Outcome} \\
\midrule
Abstention (\ref{sec:abstain}) & Accept only the answers the probe scores highest & Helps: at 70\% coverage, accuracy on the accepted answers is 92.6\% against 88.4\% for surface features alone. The earlier, answer-free probe adds nothing measurable to a baseline that knows how the finished samples agree \\
Reranking (\ref{sec:rerank}) & Return one of four finished answers & Beats random choice, but a selector that sees only surface features does about as well \\
Pruning (\ref{sec:prune}) & Decide which branches of the computation continue & Not testable on my data: the branches rarely differ in outcome. Pruning sampled answers beats random pruning but is unresolved against best-of-3 and best-of-4 \\
Steering (\ref{sec:steering}) & Add a probe-derived direction to the hidden states, chosen in seven ways & No consistent change in success \\
Branching (\ref{sec:branching}) & Perturb a forked branch and let it continue & No evidence of a gain over ordinary sampling, on four tasks \\
Loops per question (\ref{sec:depth}) & Let a probe or Ouro's exit gate choose which questions get more loops & No robust gain over random assignment of the same mix of depths \\
Light training (\ref{sec:training}) & Train a LoRA adapter to respond to a direction & No resolved benefit over an adapter trained with shuffled signs \\
\end{tabularx}
\end{floatscopebox}
\end{table}

\begin{figure}[t]
\centering
\includegraphics[width=\linewidth]{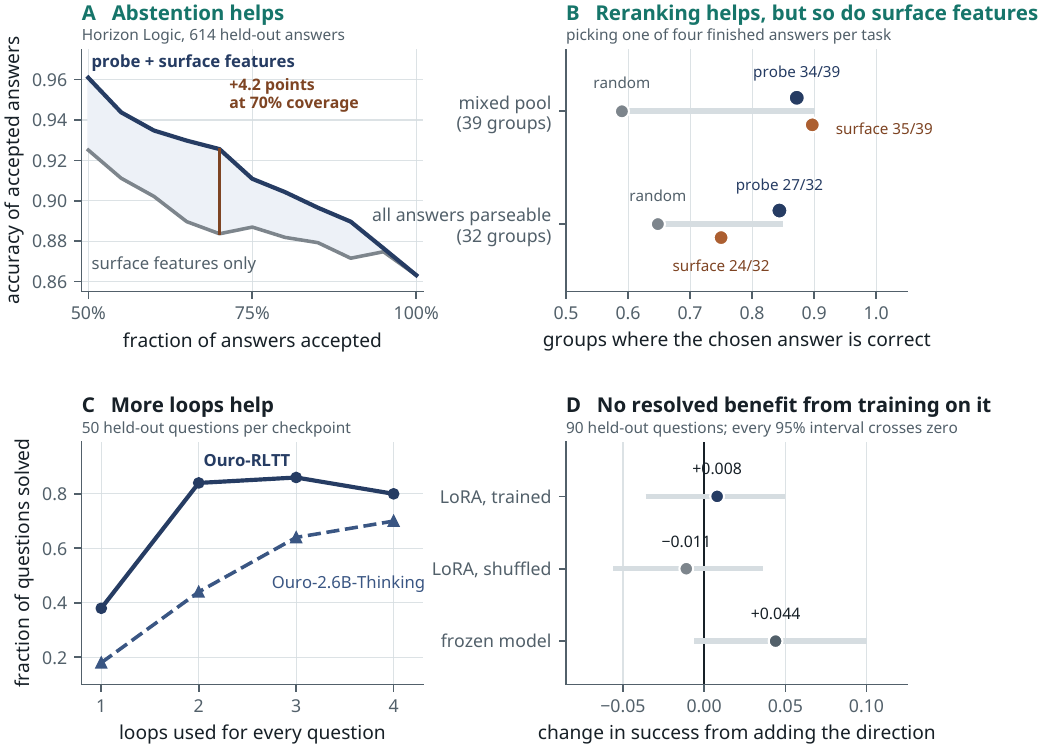}
\caption{\textbf{Using the signal} (Horizon Logic). (A)~Abstention: accepting only the answers the probe scores highest raises accuracy on the accepted answers (614 held-out answers). (B)~Reranking four finished answers: the probe beats random choice on both pools, but a selector that sees only surface features does about as well. (C)~Fixed-depth success on 50 held-out questions: more loops help. Which questions get them is tested in Section~\ref{sec:depth}. (D)~A LoRA adapter trained to respond to a direction (Section~\ref{sec:training}) shows no resolved benefit over one trained with shuffled signs.}
\label{fig:use}
\end{figure}

\subsection{Abstention}\label{sec:abstain}
Suppose the model may decline to answer. If it accepts only the answers its probe scores highest, how accurate are the answers it gives? Figure~\ref{fig:use}A shows this for the 614 held-out Horizon answers, from 201 questions, of the final-answer-line test (Section~\ref{sec:results3}). Ranking answers by the probe on hidden states plus surface features (the combined probe of Section~\ref{sec:results3}, which scores like the hidden-only one), rather than by surface features alone, raises accuracy at every recorded coverage below 100\%: at 70\% coverage it is 92.6\% against 88.4\%, though at 95\% the difference is within noise. Over all coverage, the area under the accuracy--coverage curve rises by 0.029 ($[0.014, 0.046]$); between 50 and 100\% coverage it rises by 0.0124 ($[0.0048, 0.0207]$), and that gain holds against a larger baseline that adds the number of unparseable answers to the same question (0.0119) and on the new questions alone (0.0093, $[0.0006, 0.0184]$).

The Horizon gain is small but consistent, and it does not depend on whether the conclusion was already stated: deciding whether to trust an answer once its reasoning is written is a use in its own right \citep{geifman2017selective}.

The probe above reads the reasoning up to the final-answer line. For the answer-free probe of Section~\ref{sec:fixed} I tried a stronger baseline, one that uses what the other samples say: when sampled answers agree, the common answer is usually right \citep{wang2023selfconsistency}. On the answer-free runs, each question has four sampled answers. I gave a baseline what the probe cannot know at its cut: the length, log-probability and token entropy of each finished answer, how many of the other answers agree with it, and the question features. Against this baseline the answer-free probe adds nothing measurable. For single answers, the area under the risk--coverage curve (lower is better) changes by $-0.0004$ ($[-0.0014, 0.0005]$) on the second Horizon set and $+0.0004$ ($[-0.0003, 0.0010]$) on GSM8K.

The probe is not a copy of agreement. Among Horizon answers that every other parseable answer agrees with, it still separates correct from incorrect (AUROC 0.69, $[0.60, 0.77]$; 36 incorrect of 1,855). But it carries too little beyond this baseline to help a decision that can wait for the samples to finish; because the baseline also holds length, confidence and question features, this does not isolate agreement. Its advantage is that it comes earlier. This comparison was run after the results (Appendix~\ref{app:use}).

\subsection{Reranking}\label{sec:rerank}
Now suppose I sample four answers and must return one. A correctness probe on the hidden states of each complete answer can pick the one it scores highest, as a verifier does in best-of-$N$ selection \citep{cobbe2021gsm8k}. I tested this on two pools.

In the first, the four answers were sampled freely, so some never give a parseable answer. On the 39 held-out questions whose answers include both correct and incorrect ones, the probe picks a correct answer 34 times, where random choice would expect 23.0 ($p=2.4\times10^{-5}$). But a selector that sees no hidden states at all, only the answer's length, whether it contains a final-answer line, whether it hit the token limit, and one length ratio, does as well, picking correctly 35 times. The reason is visible in the pool: 34 of the 39 groups contain an unparseable answer, and in 26 of them simply avoiding that answer guarantees a correct pick.

The second pool removes that shortcut. I resampled until each question had four parseable answers, without ever looking at correctness. On the 32 held-out groups (of 110) with both kinds, the probe picks correctly 27 times, against 20.7 expected by chance ($p=0.0086$). The surface-only selector picks correctly 24 times, and the probe's advantage over it, $+0.09$ ($[-0.06, 0.25]$), is not resolved at this sample size.

\subsection{Pruning branches inside the loop}\label{sec:prune}
A probe can also act while the computation is still running. If the model's state is forked into branches, a probe can decide which branches continue through the remaining layers and loops, so that weak ones never receive the rest of the computation.

I built this search into Ouro's loop. At layers 24, 36 and~47 on each loop, every surviving branch spawns children through a small perturbation of its hidden state; a ranking probe scores all candidates from the hidden states at that point, computed over the prompt before any branch has generated text; and at most eight continue to the next stage. The search covered 48 tasks. On the 9 of them that the probes never saw in training, the branch with the best final answer survived 96 of the 99 pruning decisions (97.0\%).

That number says less than it seems. Perturbed branches rarely change the outcome: over all 48 tasks, 92.8\% of children end with the same reward as their parent. In 35 of the 48 tasks every branch has the same reward, so any pruning keeps a best branch. Only about five tasks could separate the probe from random pruning, so these data cannot show that the probe prunes better than chance.

A later test pruned sampled answers instead, replayed offline: six answers per question were generated to completion, and the policy kept the two the probe scored highest after 80 tokens and returned the one it preferred on the full answer. For this version I reran it, sealed before generation, on 180 harder Horizon Logic questions (proof depth 4), where 83.3\% of questions have a correct answer among six. The pruning beat 95\% of random prunings at the same budget (0.456 against 0.439), so its early choices are better than chance. Against finishing three samples and letting the same probe pick one (0.406), it gained $+0.050$ ($[-0.022, 0.122]$), and against finishing four (0.422) $+0.033$ ($[-0.039, 0.100]$). Both intervals include zero as well as sizable gains, so the comparison is unresolved, not a tie. Counted as six 80-token prefixes and two continuations, the pruning would also generate more tokens than best-of-3 (868 per question against 814).

\subsection{Steering}\label{sec:steering}
The simplest way to act on a direction is to add it to the hidden states during generation and see whether behavior moves the way the probe predicts \citep{turner2023steering,panickssery2023caa,li2023iti}. I first checked that the machinery works: adding a direction with zero strength reproduces unmodified generation exactly, the effect grows with strength, and it reaches later states and the output, lasting about 32 tokens.

I then tried seven ways of choosing the direction: the probe's weight vector; the difference between the mean states of successful and unsuccessful answers; an RMS-calibrated version of the probe direction; the gradient of a probe's score; and three trained adapters, one derived from a classifier, one trained to raise the logit of the correct answer under teacher forcing, and one trained with REINFORCE on a sequence-level reward. None produced a consistent change in success on held-out questions. Outputs changed, but not in the intended direction. The REINFORCE adapter is the clearest case: it raised the reward during training, then on 12 held-out tasks did no better than a random direction of the same size. The evaluations are small (Appendix~\ref{app:steering}).

That probe directions make poor steering directions is not new. A logistic-regression probe can pick up any correlate of the label \citep{hewitt2019control,belinkov2022probing}, directions found by different methods differ in how much they causally affect the output \citep{marks2023geometry}, steering vectors are often unreliable \citep{tan2024steering}, and the Euclidean angle between a probe direction and an intervention direction need not measure how related they are \citep{park2023linear}.

\subsection{Branching}\label{sec:branching}
A looped model offers another way to act: fork the computation at some layer and loop, perturb the branch, and let it continue on its own. To test this I implemented branching over Ouro's KV cache, which holds one entry per layer and loop, 192 per token. Each branch keeps its own copy. An unperturbed branch reproduces the unforked run bit-for-bit during prefill and within bf16 rounding during decoding. Storing the residual at the fork point lets a branch recompute only the layers after it, which matches full recomputation exactly (Appendix~\ref{app:branching}).

With branching in place, the question is whether perturbed branches reach correct answers that ordinary sampling does not. On four tasks, deterministic branches produced no new correct answers. Sampled branches sometimes did, but drawing the same number of ordinary samples did better: plain sampling found a correct answer in 75\% of cases and sampled branches in 61\%. Four tasks is a small screen. It removes the evidence for a gain and says nothing more.

\subsection{Loops per question}\label{sec:depth}
Ouro solves more questions with more loops. On 180 new Horizon questions, 50 of them held out for evaluation, with two answers sampled per question and a question counted as solved if either is correct, Ouro-RLTT solves 38\% with one loop and 84\% with two (86\% with three, 80\% with four), and Ouro-2.6B-Thinking goes from 18\% with one loop to 70\% with four (Figure~\ref{fig:use}C). With a fixed budget of loops, the natural policy is to give more loops to harder questions \citep{graves2016act}.

I compared three ways of choosing which questions get which depth: a probe on the prompt's hidden states at loop~1 (layers 8, 16 and~24), trained to predict the smallest depth that suffices; Ouro's own exit gate, averaged over the prompt; and random assignment, repeated 100 times. Each random assignment puts the same number of questions at each depth as the policy it is compared with.

The probe did not beat the 95th percentile of random assignment at any of the 19 budgets on either checkpoint. The exit gate beat it once, on Ouro-RLTT at an average depth of 3.52 loops (0.88 against 0.861), which with 50 held-out questions is not a robust gain. The two did not differ from each other (differences of $-0.06$ to $+0.04$ on Ouro-RLTT, all intervals crossing zero). The mix of depths matters. Which questions receive the extra loops, as chosen by either method, makes no robust difference. This does not contradict \citet{zhu2025ouro}, whose gate beats a fixed exit depth on MMLU: there the gate chooses a depth for each token and is compared with giving every token the same depth, not with random assignment of the same mix of depths.

\subsection{Light training}\label{sec:training}
Could a little training connect the signal to control? Two small experiments suggest not, although neither is a strong test. A 300-step LoRA fine-tune (1.1\% of the parameters) changed surface behavior but not the fraction of tasks where some sample is correct (pass@$K$ 0.708 before, 0.688 after).

In the second experiment I trained a LoRA adapter so that adding a fixed direction at layer~24 on loop~1 accompanies correct answers and subtracting it accompanies incorrect ones, together with a twin adapter trained with the signs shuffled. If this binding could be learned so cheaply, adding the direction at test time should raise success. It did not: the change was $+0.008$ ($[-0.036, 0.050]$) for the trained adapter, against $-0.011$ for the shuffled one (difference $+0.019$, $[-0.044, 0.081]$; 90 held-out questions; Figure~\ref{fig:use}D), although the intervals do not rule out a gain of a few points. Both adapters also lowered plain accuracy, by 9.4 and 7.2 points. Both experiments are small: one direction, one layer, 147 training steps per adapter, trained off-policy.

\subsection{Why reading does not become control}
A tidy explanation would be geometric: perhaps the directions my interventions can write lie outside the directions that matter for success. I tested this, and the data do not support it. The direction separating correct from incorrect continuations projects into the span of my branch perturbations slightly more than a random direction would (0.018 of its energy, against 0.014 expected), but slightly less than directions fitted to shuffled labels do (mean 0.023); it rests on only 8 correct branches among 344 (Appendix~\ref{app:geometry}). The remaining candidates are less tidy: perturbations propagate nonlinearly through the loops, the magnitudes I could apply safely may be too small, and nothing in training ever tied these directions to outcomes. The last is the one I would test next.

\section{Related work}\label{sec:related}

\paragraph{Predicting correctness from hidden states} Language models can be trained to predict whether they know the answer to a question \citep{kadavath2022know}, and probes on their hidden states predict whether a statement is true \citep{azaria2023lying}, and whether an answer will be a hallucination, in some cases before any text is generated \citep{orgad2024know,kossen2024sep}. For reasoning, probes predict whether a chain of thought will succeed from the representation before it starts \citep{afzal2025knowing} or after its first few tokens \citep{david2025temporal}, and whether an intermediate answer in a reasoning trace is correct, which allows reasoning to stop early \citep{zhang2025reasoning}. \citet{afzal2025knowing} also compare their probe with a BERT reader of the generated tokens, which does worse. \citet{sun2026trajectories} find that the hidden states of correct and incorrect solutions look alike in early reasoning steps and separate in late ones, which still allows correctness to be predicted before the answer; they also steer along these trajectories with gains on long problems, unlike the probe-derived directions of Section~\ref{sec:steering}.

\citet{lindsey2025biology} trace a circuit by which a model distinguishes entities it knows from ones it does not. Like \citet{david2025temporal} and \citet{sun2026trajectories}, my answer-free test reads the computation after the reasoning has started and before any answer is stated. What it adds is a cut chosen without labels by a rule fixed in advance, baselines for question difficulty and for a reader of the same text, a comparison between answers to the same question, and the looped setting with the loop analysis of Section~\ref{sec:loops}.

\paragraph{Looped and recurrent models} Universal Transformers \citep{dehghani2019ut} apply one block repeatedly, and recent work scales the idea to language models: looped transformers \citep{saunshi2025looped}, recurrent-depth models such as Huginn \citep{geiping2025huginn}, and Ouro \citep{zhu2025ouro}, which RLTT further trains \citep{williams2026rltt}. Coconut \citep{hao2024coconut} instead feeds hidden states back as inputs along the sequence. Adaptive computation lets a model choose how many steps to take \citep{graves2016act}; Ouro learns this with an exit gate. The tuned lens \citep{belrose2023tuned} asks whether a decoder fitted at one depth works at another, and my cross-loop transfer asks the same question of probes across loops. \citet{zhu2025ouro} also probe intermediate states across loops.

\paragraph{Probing and steering} A probe shows what can be decoded from a representation, not what the model uses \citep{hewitt2019control,belinkov2022probing,elazar2021amnesic}. Steering adds directions to activations to change behavior \citep{turner2023steering,panickssery2023caa,li2023iti,zou2023repe,arditi2024refusal}, with mixed reliability \citep{tan2024steering}, and directions that decode a property are not always the ones that control it \citep{marks2023geometry,park2023linear}. My steering results add a looped model and a success target to these cases.

\paragraph{Selection and abstention} Learned verifiers select among sampled solutions \citep{cobbe2021gsm8k}, process reward models score intermediate steps \citep{lightman2023verify}, and selective classification studies when a model should abstain \citep{geifman2017selective}. Agreement among sampled answers is itself a strong signal of correctness \citep{wang2023selfconsistency}, and Section~\ref{sec:use} uses it as a baseline. My decision-level uses are instances of these, with a probe on hidden states as the scorer.

\paragraph{Introspection} A separate line of work asks whether models can report on their own internal states \citep{binder2024looking,lindsey2025introspection,comsa2025introspection,song2025privileged}. \citet{lindsey2025introspection} defines introspective awareness through accurate, causally grounded self-report and asks that ``introspection'' be reserved for a model's access to its own states. My probes read the model from outside, and I make no claim that the model accesses the information they find. The term I use, operational proto-introspection, names a narrower property: information in the hidden states that a probe can read and that points to how the computation will end. It is a precursor of introspection at most, and the prefix marks that difference.

\paragraph{Branching over KV caches} Serving systems share and fork KV caches across branches \citep{kwon2023paged,zheng2024sglang,miao2024specinfer}, and MELT shrinks Ouro's per-loop cache to a single gated one \citep{conchello2026melt}. My branching adds forks inside a loop, which requires recomputing the layers after the fork from a stored residual (Appendix~\ref{app:branching}).

\section{Discussion}\label{sec:discussion}

Looped models make a model's computation readable at more points than usual, and two things stand out in Ouro. Early in the reasoning, before any answer is stated, the hidden states already predict whether the eventual answer will be correct. Across questions, a baseline of token confidence and features of the question predicts about as well. On the logic task the signal also holds between two answers to the same question, where difficulty is fixed, and on the second set it tracks how soon an answer will commit. A fine-tuned reader of the visible text predicts about as well, so the hidden states are one convenient place to read this information, not the only one.

The information that ranks answers becomes readable in early layers mostly with the second loop. It lies along much the same direction on every loop, and the first loop seems to hold it weakly rather than in a different form. Ouro's own results also show the second loop bringing the largest gain. Any account of how looped models work should explain why one pass through the layers leaves so little of it.

The practical picture is mixed. A cheap probe on hidden states can help decide whether to trust an answer and which answer to return, although once several samples are finished, a baseline built on their agreement leaves the early probe nothing measurable to add. The ranking probes of Section~\ref{sec:loops} work across closely related checkpoints without retraining, which matters for monitoring \citep{korbak2025cot}; applied to a different target, predicting which of the model's own sampled answers are correct, they score at or below chance (AUROC 0.35--0.48) in a small test on 16 questions. And in the frozen model I found no way to turn the signal into better generation, although the directions I steered with came from older probes, not from the answer-free one.

My best guess is that nothing in training connects the directions the model's states can be pushed along to the outcomes a probe reads, and that the connection would have to be trained in. Probe scores could serve as dense rewards, anchored to an answer checker, or branch selection could be trained directly, as latent-reasoning methods train their search \citep{hao2024coconut}. Either use needs care, since a model optimized against a probe can learn to fool it. The use I find most promising, an answer-free probe that decides which branches continue inside the loop, first needs branches that differ in outcome.

\paragraph{Limitations}
\begin{itemize}
\item \textbf{One model family.} All results come from Ouro models of 2.6B parameters or fewer, with frozen weights. Huginn replicates only the rise of readability with recurrence, and MiniCPM only the existence of a ranking signal.
\item \textbf{Section~\ref{sec:correct}.} The positive answer-free results come from one task family and one checkpoint, at one cut chosen without labels, and cover answers that state no answer within 48 tokens and later give a parseable one. Across questions, a baseline with question features predicts about as well. The within-question result rests on 73 held-out pairs from 43 questions; readers of the same text match the probe there on the second set. The difficulty baseline uses the correct option, and whether the hidden states add to difficulty across questions depends on the cross-validation folds. On GSM8K and on Ouro-2.6B-Thinking the tests are unresolved.
\item \textbf{Section~\ref{sec:loops}.} One probing target, ranking candidates in five domains, where many wrong candidates are constructed corruptions, with 120 held-out groups per domain. The moves to earlier loops were added after the first results and run on Ouro-RLTT only; one probe per cell, with no correction for the 18 moves tested.
\item \textbf{Section~\ref{sec:use}.} The advantage of hidden states over surface features in reranking is unresolved. The branch-pruning probe cannot be compared with random pruning, because the branches rarely differ in outcome; pruning sampled answers, replayed offline, showed no resolved advantage over best-of-3 or best-of-4 on 180 questions. The steering and training experiments cover specific methods, magnitudes and small samples, and larger or end-to-end training was not tried.
\item \textbf{Availability.} The Ouro-RLTT weights are not public. All derived features, predictions and statistics are preserved (Appendix~\ref{app:repro}).
\end{itemize}

\paragraph{Conclusion} I probed a looped language model while it reasons. Before it states an answer, its hidden states predict whether that answer will be correct. Token confidence plus simple features of the question predict about as well across questions, but on the logic task the signal also holds between two answers to the same question, where difficulty is fixed. Information about answer quality becomes readable in early layers mostly with the second loop, along a direction largely shared across loops. These signals help decide what to do with finished answers, although surface features rerank about as well, and once several samples are finished a baseline that knows their agreement leaves the early probe nothing measurable to add. I could not use them to make the frozen model generate better answers.

\begin{kirinacknowledgment}
I thank Jonathan Williams for providing the Ouro-RLTT weights used throughout this work; without them, the RLTT results reported here would not exist. I owe Section~\ref{sec:loops} to Associate Professor Goran Đambić, PhD, of Algebra Bernays University, who asked whether the earlier layers might already be readable once the model had been round the loop a few times. I had not thought to check. They are. I thank Denis Janiak, whose comparison of correctness signals raised the question of agreement among samples that Section~\ref{sec:use} tests.
\end{kirinacknowledgment}

\renewcommand{\bibfont}{\small}
\setlength{\bibsep}{0.6pt plus 0.6pt}
\bibliographystyle{plainnat}
\bibliography{refs}

\begin{thebibliography}{46}
\providecommand{\natexlab}[1]{#1}
\providecommand{\url}[1]{\texttt{#1}}
\expandafter\ifx\csname urlstyle\endcsname\relax
  \providecommand{\doi}[1]{doi: #1}\else
  \providecommand{\doi}{doi: \begingroup \urlstyle{rm}\Url}\fi

\bibitem[Afzal et~al.(2025)Afzal, Matthes, Chechik, and
  Ziser]{afzal2025knowing}
Anum Afzal, Florian Matthes, Gal Chechik, and Yftah Ziser.
\newblock Knowing before saying: {LLM} representations encode information about
  chain-of-thought success before completion.
\newblock In \emph{Findings of the Association for Computational Linguistics:
  ACL 2025}, 2025.
\newblock arXiv:2505.24362.

\bibitem[Arditi et~al.(2024)Arditi, Obeso, Syed, Paleka, Panickssery, Gurnee,
  and Nanda]{arditi2024refusal}
Andy Arditi, Oscar Obeso, Aaquib Syed, Daniel Paleka, Nina Panickssery, Wes
  Gurnee, and Neel Nanda.
\newblock Refusal in language models is mediated by a single direction.
\newblock \emph{arXiv preprint arXiv:2406.11717}, 2024.

\bibitem[Azaria and Mitchell(2023)]{azaria2023lying}
Amos Azaria and Tom Mitchell.
\newblock The internal state of an {LLM} knows when it's lying.
\newblock In \emph{Findings of the Association for Computational Linguistics:
  EMNLP 2023}, 2023.
\newblock arXiv:2304.13734.

\bibitem[Belinkov(2022)]{belinkov2022probing}
Yonatan Belinkov.
\newblock Probing classifiers: Promises, shortcomings, and advances.
\newblock \emph{Computational Linguistics}, 48\penalty0 (1):\penalty0 207--219,
  2022.
\newblock arXiv:2102.12452.

\bibitem[Belrose et~al.(2023)Belrose, Furman, Smith, Halawi, Ostrovsky,
  McKinney, Biderman, and Steinhardt]{belrose2023tuned}
Nora Belrose, Zach Furman, Logan Smith, Danny Halawi, Igor Ostrovsky, Lev
  McKinney, Stella Biderman, and Jacob Steinhardt.
\newblock Eliciting latent predictions from transformers with the tuned lens.
\newblock \emph{arXiv preprint arXiv:2303.08112}, 2023.

\bibitem[Binder et~al.(2024)Binder, Chua, Korbak, et~al.]{binder2024looking}
Felix~J. Binder, James Chua, Tomek Korbak, et~al.
\newblock Looking inward: Language models can learn about themselves by
  introspection.
\newblock \emph{arXiv preprint arXiv:2410.13787}, 2024.

\bibitem[Cobbe et~al.(2021)Cobbe, Kosaraju, Bavarian, et~al.]{cobbe2021gsm8k}
Karl Cobbe, Vineet Kosaraju, Mohammad Bavarian, et~al.
\newblock Training verifiers to solve math word problems.
\newblock \emph{arXiv preprint arXiv:2110.14168}, 2021.

\bibitem[Com{\c{s}}a and Shanahan(2025)]{comsa2025introspection}
Iulia~M. Com{\c{s}}a and Murray Shanahan.
\newblock Does it make sense to speak of introspection in large language
  models?
\newblock \emph{arXiv preprint arXiv:2506.05068}, 2025.

\bibitem[Conchello~Vendrell et~al.(2026)Conchello~Vendrell, Padres~Masdemont,
  Grillo, Ros-Giralt, Behboodi, and Massoli]{conchello2026melt}
Victor Conchello~Vendrell, Arnau Padres~Masdemont, Niccol{\`o} Grillo, Jordi
  Ros-Giralt, Arash Behboodi, and Fabio~Valerio Massoli.
\newblock Memory-efficient looped transformer: Decoupling compute from memory
  in looped language models.
\newblock \emph{arXiv preprint arXiv:2605.07721}, 2026.

\bibitem[David(2025)]{david2025temporal}
Joey David.
\newblock Temporal predictors of outcome in reasoning language models.
\newblock \emph{arXiv preprint arXiv:2511.14773}, 2025.

\bibitem[Dehghani et~al.(2019)Dehghani, Gouws, Vinyals, Uszkoreit, and
  Kaiser]{dehghani2019ut}
Mostafa Dehghani, Stephan Gouws, Oriol Vinyals, Jakob Uszkoreit, and {\L}ukasz
  Kaiser.
\newblock Universal transformers.
\newblock In \emph{International Conference on Learning Representations}, 2019.
\newblock arXiv:1807.03819.

\bibitem[Elazar et~al.(2021)Elazar, Ravfogel, Jacovi, and
  Goldberg]{elazar2021amnesic}
Yanai Elazar, Shauli Ravfogel, Alon Jacovi, and Yoav Goldberg.
\newblock Amnesic probing: Behavioral explanation with amnesic counterfactuals.
\newblock \emph{Transactions of the Association for Computational Linguistics},
  9:\penalty0 160--175, 2021.
\newblock arXiv:2006.00995.

\bibitem[Geifman and El-Yaniv(2017)]{geifman2017selective}
Yonatan Geifman and Ran El-Yaniv.
\newblock Selective classification for deep neural networks.
\newblock In \emph{Advances in Neural Information Processing Systems}, 2017.
\newblock arXiv:1705.08500.

\bibitem[Geiping et~al.(2025)Geiping, McLeish, Jain, Kirchenbauer, Singh,
  Bartoldson, Kailkhura, Bhatele, and Goldstein]{geiping2025huginn}
Jonas Geiping, Sean McLeish, Neel Jain, John Kirchenbauer, Siddharth Singh,
  Brian~R. Bartoldson, Bhavya Kailkhura, Abhinav Bhatele, and Tom Goldstein.
\newblock Scaling up test-time compute with latent reasoning: A recurrent depth
  approach.
\newblock \emph{arXiv preprint arXiv:2502.05171}, 2025.

\bibitem[Graves(2016)]{graves2016act}
Alex Graves.
\newblock Adaptive computation time for recurrent neural networks.
\newblock \emph{arXiv preprint arXiv:1603.08983}, 2016.

\bibitem[Hao et~al.(2024)Hao, Sukhbaatar, Su, Li, Hu, Weston, and
  Tian]{hao2024coconut}
Shibo Hao, Sainbayar Sukhbaatar, DiJia Su, Xian Li, Zhiting Hu, Jason Weston,
  and Yuandong Tian.
\newblock Training large language models to reason in a continuous latent
  space.
\newblock \emph{arXiv preprint arXiv:2412.06769}, 2024.

\bibitem[Hewitt and Liang(2019)]{hewitt2019control}
John Hewitt and Percy Liang.
\newblock Designing and interpreting probes with control tasks.
\newblock In \emph{Proceedings of EMNLP-IJCNLP}, 2019.
\newblock arXiv:1909.03368.

\bibitem[Hu et~al.(2024)Hu, Tu, et~al.]{hu2024minicpm}
Shengding Hu, Yuge Tu, et~al.
\newblock {MiniCPM}: Unveiling the potential of small language models with
  scalable training strategies.
\newblock \emph{arXiv preprint arXiv:2404.06395}, 2024.

\bibitem[Kadavath et~al.(2022)Kadavath, Conerly, Askell,
  et~al.]{kadavath2022know}
Saurav Kadavath, Tom Conerly, Amanda Askell, et~al.
\newblock Language models (mostly) know what they know.
\newblock \emph{arXiv preprint arXiv:2207.05221}, 2022.

\bibitem[Kapoor and Narayanan(2023)]{kapoor2023leakage}
Sayash Kapoor and Arvind Narayanan.
\newblock Leakage and the reproducibility crisis in machine-learning-based
  science.
\newblock \emph{Patterns}, 4\penalty0 (9), 2023.
\newblock arXiv:2207.07048.

\bibitem[Kirin(2026)]{kirin2026rpe}
Jan Kirin.
\newblock Reading human preference from looped transformer internal states.
\newblock \emph{arXiv preprint arXiv:2604.09870}, 2026.
\newblock Versions 1 and 2, titled \emph{Relational Preference Encoding in
  Looped Transformer Internal States}, carry the original claims and the
  erratum; version 3 is the corrected paper.

\bibitem[Korbak et~al.(2025)Korbak, Balesni, Barnes, et~al.]{korbak2025cot}
Tomek Korbak, Mikita Balesni, Elizabeth Barnes, et~al.
\newblock Chain of thought monitorability: A new and fragile opportunity for
  {AI} safety.
\newblock \emph{arXiv preprint arXiv:2507.11473}, 2025.

\bibitem[Kossen et~al.(2024)Kossen, Han, Razzak, Schut, Malik, and
  Gal]{kossen2024sep}
Jannik Kossen, Jiatong Han, Muhammed Razzak, Lisa Schut, Shreshth Malik, and
  Yarin Gal.
\newblock Semantic entropy probes: Robust and cheap hallucination detection in
  {LLMs}.
\newblock \emph{arXiv preprint arXiv:2406.15927}, 2024.

\bibitem[Kwon et~al.(2023)Kwon, Li, Zhuang, et~al.]{kwon2023paged}
Woosuk Kwon, Zhuohan Li, Siyuan Zhuang, et~al.
\newblock Efficient memory management for large language model serving with
  {PagedAttention}.
\newblock In \emph{Proceedings of the 29th Symposium on Operating Systems
  Principles}, 2023.
\newblock arXiv:2309.06180.

\bibitem[Li et~al.(2023)Li, Patel, Vi{\'e}gas, Pfister, and
  Wattenberg]{li2023iti}
Kenneth Li, Oam Patel, Fernanda Vi{\'e}gas, Hanspeter Pfister, and Martin
  Wattenberg.
\newblock Inference-time intervention: Eliciting truthful answers from a
  language model.
\newblock In \emph{Advances in Neural Information Processing Systems}, 2023.
\newblock arXiv:2306.03341.

\bibitem[Lightman et~al.(2023)Lightman, Kosaraju, Burda,
  et~al.]{lightman2023verify}
Hunter Lightman, Vineet Kosaraju, Yura Burda, et~al.
\newblock Let's verify step by step.
\newblock \emph{arXiv preprint arXiv:2305.20050}, 2023.

\bibitem[Lindsey(2025)]{lindsey2025introspection}
Jack Lindsey.
\newblock Emergent introspective awareness in large language models.
\newblock \emph{Transformer Circuits Thread}, 2025.
\newblock arXiv:2601.01828.

\bibitem[Lindsey et~al.(2025)Lindsey, Gurnee, Ameisen, Chen, Pearce, Turner,
  Citro, et~al.]{lindsey2025biology}
Jack Lindsey, Wes Gurnee, Emmanuel Ameisen, Brian Chen, Adam Pearce,
  Nicholas~L. Turner, Craig Citro, et~al.
\newblock On the biology of a large language model.
\newblock \emph{Transformer Circuits Thread}, 2025.
\newblock
  \url{https://transformer-circuits.pub/2025/attribution-graphs/biology.html}.

\bibitem[Marks and Tegmark(2023)]{marks2023geometry}
Samuel Marks and Max Tegmark.
\newblock The geometry of truth: Emergent linear structure in large language
  model representations of true/false datasets.
\newblock \emph{arXiv preprint arXiv:2310.06824}, 2023.

\bibitem[Miao et~al.(2024)Miao, Oliaro, Zhang, et~al.]{miao2024specinfer}
Xupeng Miao, Gabriele Oliaro, Zhihao Zhang, et~al.
\newblock {SpecInfer}: Accelerating generative large language model serving
  with tree-based speculative inference and verification.
\newblock In \emph{Proceedings of ASPLOS}, 2024.
\newblock arXiv:2305.09781.

\bibitem[Orgad et~al.(2025)Orgad, Toker, Gekhman, Reichart, Szpektor, Kotek,
  and Belinkov]{orgad2024know}
Hadas Orgad, Michael Toker, Zorik Gekhman, Roi Reichart, Idan Szpektor, Hadas
  Kotek, and Yonatan Belinkov.
\newblock {LLMs} know more than they show: On the intrinsic representation of
  {LLM} hallucinations.
\newblock In \emph{International Conference on Learning Representations}, 2025.
\newblock arXiv:2410.02707.

\bibitem[Park et~al.(2024)Park, Choe, and Veitch]{park2023linear}
Kiho Park, Yo~Joong Choe, and Victor Veitch.
\newblock The linear representation hypothesis and the geometry of large
  language models.
\newblock In \emph{International Conference on Machine Learning}, 2024.
\newblock arXiv:2311.03658.

\bibitem[Rimsky et~al.(2024)Rimsky, Gabrieli, Schulz, Tong, Hubinger, and
  Turner]{panickssery2023caa}
Nina Rimsky, Nick Gabrieli, Julian Schulz, Meg Tong, Evan Hubinger, and
  Alexander Turner.
\newblock Steering {Llama} 2 via contrastive activation addition.
\newblock In \emph{Proceedings of the 62nd Annual Meeting of the Association
  for Computational Linguistics (Volume 1: Long Papers)}, pages 15504--15522,
  2024.
\newblock arXiv:2312.06681.

\bibitem[Saunshi et~al.(2025)Saunshi, Dikkala, Li, Kumar, and
  Reddi]{saunshi2025looped}
Nikunj Saunshi, Nishanth Dikkala, Zhiyuan Li, Sanjiv Kumar, and Sashank~J.
  Reddi.
\newblock Reasoning with latent thoughts: On the power of looped transformers.
\newblock \emph{arXiv preprint arXiv:2502.17416}, 2025.

\bibitem[Song et~al.(2025)Song, Lederman, Hu, and Mahowald]{song2025privileged}
Siyuan Song, Harvey Lederman, Jennifer Hu, and Kyle Mahowald.
\newblock Privileged self-access matters for introspection in {AI}.
\newblock \emph{arXiv preprint arXiv:2508.14802}, 2025.

\bibitem[Sun et~al.(2026)Sun, Dong, Qiao, Lin, Zhang, and
  Rajmohan]{sun2026trajectories}
Lihao Sun, Hang Dong, Bo~Qiao, Qingwei Lin, Dongmei Zhang, and Saravan
  Rajmohan.
\newblock {LLM} reasoning as trajectories: Step-specific representation
  geometry and correctness signals.
\newblock \emph{arXiv preprint arXiv:2604.05655}, 2026.

\bibitem[Tan et~al.(2024)Tan, Chanin, Lynch, Paige, Kanoulas, Garriga-Alonso,
  and Kirk]{tan2024steering}
Daniel Tan, David Chanin, Aengus Lynch, Brooks Paige, Dimitrios Kanoulas,
  Adri{\`a} Garriga-Alonso, and Robert Kirk.
\newblock Analysing the generalisation and reliability of steering vectors.
\newblock In \emph{Advances in Neural Information Processing Systems}, 2024.
\newblock arXiv:2407.12404.

\bibitem[Turner et~al.(2023)Turner, Thiergart, Leech, Udell, Vazquez, Mini, and
  MacDiarmid]{turner2023steering}
Alexander~Matt Turner, Lisa Thiergart, Gavin Leech, David Udell, Juan~J.
  Vazquez, Ulisse Mini, and Monte MacDiarmid.
\newblock Steering language models with activation engineering.
\newblock \emph{arXiv preprint arXiv:2308.10248}, 2023.

\bibitem[Wang et~al.(2024)Wang, Li, Chen, et~al.]{wang2023fair}
Peiyi Wang, Lei Li, Liang Chen, et~al.
\newblock Large language models are not fair evaluators.
\newblock In \emph{Proceedings of ACL}, 2024.
\newblock arXiv:2305.17926.

\bibitem[Wang et~al.(2023)Wang, Wei, Schuurmans, Le, Chi, Narang, Chowdhery,
  and Zhou]{wang2023selfconsistency}
Xuezhi Wang, Jason Wei, Dale Schuurmans, Quoc Le, Ed~Chi, Sharan Narang,
  Aakanksha Chowdhery, and Denny Zhou.
\newblock Self-consistency improves chain of thought reasoning in language
  models.
\newblock In \emph{International Conference on Learning Representations}, 2023.
\newblock arXiv:2203.11171.

\bibitem[Williams and Tureci(2026)]{williams2026rltt}
Jonathan Williams and Esin Tureci.
\newblock Prioritize the process, not just the outcome: Rewarding latent
  thought trajectories improves reasoning in looped language models.
\newblock \emph{arXiv preprint arXiv:2602.10520}, 2026.

\bibitem[Zhang et~al.(2025)Zhang, Chen, Pan, Zhao, Panda, Li, and
  He]{zhang2025reasoning}
Anqi Zhang, Yulin Chen, Jane Pan, Chen Zhao, Aurojit Panda, Jinyang Li, and
  He~He.
\newblock Reasoning models know when they're right: Probing hidden states for
  self-verification.
\newblock \emph{arXiv preprint arXiv:2504.05419}, 2025.

\bibitem[Zheng et~al.(2023)Zheng, Chiang, Sheng, et~al.]{zheng2023judge}
Lianmin Zheng, Wei-Lin Chiang, Ying Sheng, et~al.
\newblock Judging {LLM}-as-a-judge with {MT-Bench} and {Chatbot Arena}.
\newblock In \emph{Advances in Neural Information Processing Systems, Datasets
  and Benchmarks Track}, 2023.
\newblock arXiv:2306.05685.

\bibitem[Zheng et~al.(2024)Zheng, Yin, Xie, et~al.]{zheng2024sglang}
Lianmin Zheng, Liangsheng Yin, Zhiqiang Xie, et~al.
\newblock {SGLang}: Efficient execution of structured language model programs.
\newblock \emph{arXiv preprint arXiv:2312.07104}, 2024.

\bibitem[Zhu et~al.(2025)Zhu, Wang, Hua, Zhang, Li, et~al.]{zhu2025ouro}
Rui-Jie Zhu, Zixuan Wang, Kai Hua, Tianyu Zhang, Ziniu Li, et~al.
\newblock Scaling latent reasoning via looped language models.
\newblock \emph{arXiv preprint arXiv:2510.25741}, 2025.

\bibitem[Zou et~al.(2023)Zou, Phan, Chen, et~al.]{zou2023repe}
Andy Zou, Long Phan, Sarah Chen, et~al.
\newblock Representation engineering: A top-down approach to {AI} transparency.
\newblock \emph{arXiv preprint arXiv:2310.01405}, 2023.

\end{thebibliography}

\bigskip
\appendix
\titleformat{\section}{\sffamily\bfseries\color{KirinInk}\fontsize{13.2pt}{15.2pt}\selectfont}{\thesection}{0.6em}{}
\titleformat{\subsection}{\sffamily\bfseries\color{KirinInk}\fontsize{10.9pt}{12.8pt}\selectfont}{\thesubsection}{0.5em}{}

\begingroup\small
\section{Features and probes}\label{app:features}
Hidden states come from one frozen forward pass per text (bf16 weights, fp32 features, the library's default scaled-dot-product attention, all four loops, no early exit), averaged over valid token positions. The canonical features use layers 24, 36 and~47 on all four loops. These layers were chosen in earlier work on this model from final-loop results, so they record where the signal sits after refinement, not where it first appears; Section~\ref{sec:loops} adds layers 8 and~16.

Correctness probes standardize the features, reduce them with PCA and fit an L2-regularized logistic regression, choosing the number of components and the regularization strength by five-fold cross-validation grouped by question, on training and validation questions only. The second set of Section~\ref{sec:fixed} uses a fixed combined probe: PCA on each block of hidden states only, the surface features appended unreduced, then L2-regularized logistic regression.

Ranking probes in Section~\ref{sec:loops} use the bias-free score $w^\top(h_A-h_B)$ on unnormalized features, trained on all pairs within a group with 60 full-batch Adam steps, with a grid of 36 settings (training subsets, learning rates and seeds) selected on validation top-1 accuracy. Chance for top-1 accounts for ties within groups (0.2818). Intervals come from a question-clustered bootstrap with 10,000 draws; paired comparisons use identical draws. In 5--7 of the 600 held-out groups per cell, several candidates tie for the probe's top score because their features are identical; the pick then earns the share of them that are best, which is 0.006--0.008 lower than crediting the first one listed and changes no interval's conclusion. Every top-1 value in this paper uses this rule, including the length rankings and MiniCPM, whose groups tie more often. The one exception is the extraction check of Section~\ref{sec:loops} (0.687 and 0.685), where both values credit the first candidate listed.

Token log-probabilities in the Horizon runs and in every answer-free run are those of the sampled tokens under the sampling distribution (temperature 0.7, top-$k$ 50, top-$p$ 0.95). The GSM8K final-answer-line test used teacher-forced log-probabilities and entropies at temperature~1.

\section{Details for Section~\ref{sec:correct}}\label{app:correct}

\paragraph{GSM8K} 170 questions, 4 sampled solutions each (680 answers; 407 correct, 273 incorrect). The length features are the question's characters, the prompt's tokens and the generated tokens before the cut; the confidence features are the mean token log-probability, the mean token entropy and the entropy at the last position. The increase of 0.066 is computed from out-of-fold predictions grouped by question, with 10,000 question-clustered bootstrap draws. The text read ends at the final-answer marker or just before the first standalone appearance of the reference number, whichever comes first (Section~\ref{sec:cut}). On 600 other test problems, 747 of the 2,400 answers are right but labeled wrong by the original protocol.

\paragraph{Horizon Logic, final-answer line} The prompt is ``Answer the multiple-choice question. Give concise reasoning if useful, then end with exactly one line: FINAL ANSWER: $\langle$letter$\rangle$.'' followed by the question and the options A.~True, B.~False, C.~Unknown. The cut is the text before the first case-insensitive match of \texttt{FINAL\textbackslash s*ANSWE}. On the first 170 questions, 25.0\% of answers are unparseable and 89.0\% of the parseable ones correct; against five surface features (the four of Section~\ref{sec:results3} plus the number of unparseable answers to the same question) the increase is $+0.122$ ($[-0.014, 0.270]$) there and $+0.107$ ($[0.053, 0.164]$) pooled over 680. The hidden states also predict which answers will be unparseable (AUROC 0.713). On the 510 new questions, the check of Section~\ref{sec:results3} gives $+0.093$ ($[0.041, 0.150]$). A probe on the prompt alone, averaged over its tokens, differs from prompt length and proof depth by $+0.097$ ($[-0.006, 0.199]$) on the 510 questions and by $-0.038$ ($[-0.111, 0.038]$) on the 599 fresh ones; these are the hidden-only probe minus the baseline.

\paragraph{Answer-free test, first set}
\begin{itemize}
\item \emph{Generation.} 600 questions at positions 2,400--2,999 of the same hash-ordered pool, which no earlier run used, with the published generator's prompt, sampling and seed rule. One question was lost to a GPU memory error. Features for every cut come from one forward pass over the whole generation; because every loop of Ouro is causal, this equals a separate pass on the truncated tokens, which 24 checks confirmed exactly.
\item \emph{Composition.} 2,396 answers, 22.7\% unparseable and 26.5\% at the token limit, with 87.7\% of parseable answers correct; 309 training, 120 validation and 170 held-out questions. Answer-free share at 48 tokens: 0.757 (training) and 0.787 (held out). Increase at 16, 32, 64, 96 and 128 tokens: $+0.201$, $+0.253$, $+0.192$, $+0.156$ and $+0.126$ ($[-0.017, 0.276]$).
\item \emph{Probes.} The primary probe selected 16 principal components with $\lambda=1.0$. The TF-IDF reader combined word 1--2-grams and character 3--5-grams with logistic regression ($C=10$).
\item \emph{Which option.} A probe trained to predict the option the answer will give, applied to the 155 answers that turned out wrong, picks the correct option 69.7\% of the time ($[61.6, 77.6]$) and the answer's own option 17.4\% ($[10.8, 24.7]$). On pairs of answers to the same question that end in different options, the same probe follows each answer's own option 54\% of the time ($[47.4, 61.3]$; 214 pairs from 110 questions), not distinguishable from chance.
\item \emph{Checks after the result.} Labeling by the first final answer flips 77 of the 1,745 labels and leaves 29 incorrect held-out answers, one below the minimum of 30 fixed in advance. Without the 15 answers in the missed format (2 held out, both correct), the hidden states gain $+0.131$ ($[0.047, 0.221]$) over the TF-IDF reader, and within questions the probe gives 0.655 against the reader's 0.611 ($+0.044$, $[-0.035, 0.124]$). End-of-prompt state in the difficulty baseline: $+0.056$ ($[-0.011, 0.128]$).
\end{itemize}

\paragraph{Answer-free test, second set}
\begin{itemize}
\item \emph{Generation and labels.} Every unused proof-depth-3 and depth-4 question of the task family (1,097 and 67). Labels: whichever comes first of a final-answer marker followed by an option letter and a boxed letter; failing both, the first marker followed by an option word. In this run an option word after a marker occurs in one answer, after a letter. The run used cloud GPUs; its truncation check compares passes of equal length (maximum difference 0).
\item \emph{Composition.} 4,656 answers, 70.4\% labeled, 36.3\% at the 448-token limit, 89.9\% of labeled answers correct; the cut is 48 tokens (76.1\% of training answers state no answer). Increases at 16, 32, 64, 96 and 128 tokens: $+0.018$, $+0.027$, $+0.052$, $+0.038$ and $+0.093$; held-out within-question AUROC 0.631, 0.714, 0.714, 0.679 and 0.607.
\item \emph{Folds.} The sealed code splits folds with scikit-learn's \texttt{GroupKFold}, whose order among groups of equal size is not fixed by a seed. Over 31 random tie orders, the end-of-prompt control ranges from 0.038 to 0.056 (clear of zero in 22) and the within-question AUROC from 0.726 to 0.740; the headline rule would read ``beyond difficulty'' in 12.
\item \emph{Within questions, out of fold.} Over all 237 pairs: probe 0.709, combined probe 0.726, baseline 0.570, TF-IDF reader 0.759 ($[0.699, 0.815]$); commit timing alone 0.781, and the probe scores 0.762 on the 185 pairs where the incorrect answer commits first. Commit timing is the position of the final-answer label in the whole answer, beyond the 48-token window.
\item \emph{Detector miss.} 9 of the 2,376 answers in the primary cohort (3 held out) state an answer in the missed format. On one alternative fold order, excluding them gives $+0.051$ ($[-0.010, 0.108]$) across questions and 0.721 ($[0.615, 0.816]$; 68 pairs) within.
\end{itemize}

\paragraph{Out-of-fold procedures} The within-question values 0.709, 0.738 and 0.717 for the second set come from different fits: nested out-of-fold scoring; the reader comparisons, which select each model's settings once on training and validation questions and then score every fold with them, so that their all-splits comparisons are descriptive; and the hash folds of the second-checkpoint analysis. The held-out comparisons are not affected.

\paragraph{Fine-tuned text readers (added after the results)} The readers see the prompt and the first 48 generated tokens. DeBERTa-v3-large, fully fine-tuned, and Qwen2.5-1.5B with a rank-16 LoRA and a classification head pick their learning rate (three values) and number of epochs (two to four) by grouped cross-validation on training and validation questions, with one seed; every choice was hashed before the first fit. First set, within questions on the 57 held-out pairs (probe 0.719): DeBERTa 0.754, Qwen 0.456; across questions under first-final-answer labels (probe 0.777): DeBERTa 0.523, Qwen 0.617. Second set, across questions (probe 0.683): DeBERTa 0.701, Qwen 0.651; within on the 73 held-out pairs (probe 0.740): DeBERTa 0.719 ($+0.021$, $[-0.087, 0.121]$), Qwen 0.616; out of fold over 237 pairs (probe 0.738): DeBERTa 0.738, Qwen 0.643.

\paragraph{GSM8K at a fixed cut} 600 test problems, labeled by the first number the answer commits to (after a final-answer marker, else after ``\#\#\#\#'', else after ``the answer is'', else in a box); taking the earliest match of any format instead changes 5 of 1,863 labels here and 4 of 6,514 in the training-split run, and no result. The cut is 96 tokens. At cuts of 16 to 64 tokens (33 to 39 incorrect) the increases are $+0.014$ to $+0.034$, with every interval spanning zero.

\paragraph{GSM8K training split} The first 2,000 training problems in hash order; a batch stops once all four answers have committed to a number and written 8 more tokens. 8,000 answers, 97.8\% parseable, 94.6\% of parseable answers correct. The cut, chosen without labels, is 96 tokens (84.5\% of training answers state no answer); a hand audit of 20 prefixes found no stated answer. End-of-prompt control $+0.034$ ($[-0.018, 0.088]$). Within questions on held-out pairs, 0.482 ($[0.310, 0.655]$; 56 pairs from 20 problems); out of fold, combined 0.522, baseline 0.494, TF-IDF reader 0.447. The prompt's last state adds $+0.012$ ($[-0.093, 0.122]$) to the problem features; on the 600 test problems it adds $+0.167$ ($[0.042, 0.285]$) to a weak baseline (AUROC 0.445). On the second Horizon set it adds $+0.026$ ($[-0.010, 0.058]$) to the question features.

\paragraph{Ouro-2.6B-Thinking, answer-free test} All 4,656 prompts and seeds were checked identical to the Ouro-RLTT run. 47.2\% of answers labeled, 62.5\% at the 448-token limit, 89.2\% of labeled answers correct. The answer-free share is 77.2\% at 32 tokens and 73.2\% at 48, so the cut is 32. The plan's rule calls the result unresolved: at least 60 mixed questions, lower bound not above 0.5, upper bound not below 0.55. At 48 tokens the within-question AUROC is 0.575 ($[0.460, 0.695]$). The comparison with Ouro-RLTT pairs the two sides' bootstrap replicates in draw order; the run's sealed script paired them after sorting, which narrows the interval to $[-0.191, -0.104]$ (Ouro-2.6B-Thinking minus Ouro-RLTT). With both at 48 tokens the difference, stated the same way, is $-0.143$ ($[-0.272, -0.006]$). Generation used all four loops with the exit threshold at~1; rerun with teacher forcing, no token in 200 answers per checkpoint came from an earlier loop than the fourth. After the first three of twelve shards, the provider restarted the container with a newer GPU driver; the lost partial shards were regenerated from their seeds.

\section{Details for Section~\ref{sec:loops}}\label{app:loops}

\paragraph{Data} The five-domain candidate set with 250 training, 60 validation and 120 held-out groups per domain (8,592 candidates), split by question with no crossings; 195 math items were excluded because their IDs, built from row indices, collided across subject configurations of the dataset and so spanned splits. One forward pass per candidate, with states stored at layers 8, 16, 24, 36 and~47 on all four loops.

\begin{table}[h]
\centering
\caption{Held-out top-1 accuracy of ranking probes on Ouro-RLTT, with 95\% intervals (chance 0.282). Layers 36 and~47 were probed at loop~4 only.}
\label{tab:grid}
\small
\begin{tabular}{@{}lllll@{}}
\toprule
Layer & Loop 1 & Loop 2 & Loop 3 & Loop 4 \\
\midrule
8 & 0.403 [0.367, 0.440] & 0.561 [0.523, 0.596] & 0.627 [0.590, 0.663] & 0.614 [0.577, 0.651] \\
16 & 0.432 [0.395, 0.469] & 0.576 [0.539, 0.613] & 0.610 [0.573, 0.646] & 0.621 [0.584, 0.657] \\
24 & 0.422 [0.386, 0.458] & 0.563 [0.525, 0.599] & 0.602 [0.565, 0.639] & 0.611 [0.573, 0.648] \\
36 & & & & 0.639 [0.602, 0.675] \\
47 & & & & 0.612 [0.574, 0.649] \\
\bottomrule
\end{tabular}
\end{table}

The rise from loop~1 to loop~4 is 0.211 ($[0.163, 0.260]$) at layer~8, 0.188 ($[0.145, 0.235]$) at layer~16 and 0.189 ($[0.144, 0.234]$) at layer~24. Layer~8 at loop~3 differs from layer~24 at loop~4 by $+0.016$ ($[-0.020, 0.052]$) and from layer~36 at loop~4 by $-0.012$ ($[-0.046, 0.022]$). Before the run I fixed a margin: a cell counts as matching the late layers if it is above chance and within 0.03 of layer~24 at loop~4. Layers 8 and~16 at loops 3 and~4 meet it; at loop~2 they fall 0.034--0.050 short.

\begin{table}[h]
\centering
\caption{Moving a probe, unchanged, to another loop of the same layer (Ouro-RLTT). Each entry is the change in top-1 relative to a probe trained at the target loop, with the Spearman correlation between the two probes' scores. Every change from a loop-1 probe to a later loop has an interval excluding zero, as do the two layer-8 moves from loops 3 and~4 to loop~1 ($[-0.108, -0.025]$ and $[-0.085, -0.003]$); no other does.}
\label{tab:transfer}
\small
\begin{tabular}{@{}lllllll@{}}
\toprule
Layer & 1$\to$2 & 1$\to$3 & 1$\to$4 & 2$\to$3 & 2$\to$4 & 3$\to$4 \\
\midrule
8 & $-0.102$ (0.69) & $-0.234$ (0.55) & $-0.211$ (0.53) & $-0.023$ (0.90) & $+0.007$ (0.94) & $+0.008$ (0.91) \\
16 & $-0.112$ (0.85) & $-0.135$ (0.77) & $-0.131$ (0.78) & $-0.005$ (0.95) & $+0.001$ (0.95) & $+0.003$ (0.99) \\
24 & $-0.085$ (0.85) & $-0.087$ (0.77) & $-0.108$ (0.77) & $+0.001$ (0.92) & $-0.008$ (0.90) & $+0.018$ (0.95) \\
\midrule
 & 2$\to$1 & 3$\to$1 & 4$\to$1 & 3$\to$2 & 4$\to$2 & 4$\to$3 \\
\midrule
8 & $-0.017$ (0.74) & $-0.067$ (0.55) & $-0.045$ (0.60) & $-0.010$ (0.87) & $+0.002$ (0.94) & $-0.015$ (0.91) \\
16 & $-0.004$ (0.86) & $-0.008$ (0.80) & $-0.021$ (0.79) & $-0.008$ (0.95) & $-0.005$ (0.93) & $-0.005$ (0.98) \\
24 & $+0.003$ (0.88) & $+0.013$ (0.78) & $+0.009$ (0.77) & $-0.012$ (0.92) & $-0.023$ (0.89) & $+0.008$ (0.96) \\
\bottomrule
\end{tabular}
\end{table}

Before scoring at the target loop, each moved probe was checked to reproduce its stored score at the source loop, and its weights were checked unchanged by hash. Sign agreement between moved and target probes on pairs of candidates is 0.71--0.87 for probes from loop~1 and 0.92--0.97 for the others. For the moves to earlier loops, the saved weights first reproduced every stored prediction file and every stored transfer interval exactly. Cosine similarity between the weight vectors of the loop-1 and loop-4 probes is 0.27, 0.55 and 0.54 at layers 8, 16 and~24, against 0.82, 0.96 and 0.92 between loops 3 and~4. A poorly estimated probe has a low cosine with a well-estimated one even when both read the same direction, so these numbers do not separate the two explanations; the moves to loop~1 do.

\paragraph{Checks} Shuffled-label controls, one shuffled fit each, at four cells: 0.347 (layer~8, loop~1), 0.421 (layer~8, loop~4), 0.396 (layer~16, loop~2) and 0.331 (layer~24, loop~4). Ranking by text length in characters gives 0.261 with the longest first and 0.232 with the shortest first; lengths tie at the top in 117 and 152 of the 600 groups. Re-extracting layers 24, 36 and~47 along the new path on 2,434 held-out candidates gives mean cosine similarity 0.998--0.999 with the stored features. A test suite of 20 checks (scoring equivalence, label alignment, split integrity, unchanged probe weights, deterministic refits, validation-only selection and bootstrap clustering) passes.

\begin{table}[h]
\centering
\caption{Held-out top-1 accuracy on other checkpoints, same protocol as Table~\ref{tab:grid}. Ouro-1.4B has 24 layers, so its layers were mapped proportionally to 4, 8, 12, 18 and~23.}
\label{tab:family}
\small
\begin{tabular}{@{}lllllll@{}}
\toprule
Model & Layer & Loop 1 & Loop 2 & Loop 3 & Loop 4 & \\
\midrule
Ouro-2.6B (base) & 8 & 0.407 & 0.546 & 0.601 & 0.622 & \\
 & 16 & 0.423 & 0.561 & 0.626 & 0.641 & \\
 & 24 & 0.413 & 0.560 & 0.602 & 0.629 & 36: 0.632, 47: 0.581 \\
Ouro-2.6B-Thinking & 8 & 0.401 & 0.559 & 0.617 & 0.606 & \\
 & 16 & 0.414 & 0.571 & 0.611 & 0.616 & \\
 & 24 & 0.417 & 0.557 & 0.606 & 0.619 & 36: 0.624, 47: 0.601 \\
Ouro-1.4B & 4 & 0.458 & 0.517 & 0.594 & 0.596 & \\
 & 8 & 0.455 & 0.547 & 0.602 & 0.601 & \\
 & 12 & 0.466 & 0.571 & 0.616 & 0.602 & 18: 0.562, 23: 0.552 \\
\bottomrule
\end{tabular}
\end{table}

\paragraph{Huginn-0125} One forward pass per candidate with eight recurrent steps, reading the output of the last block of the recurrent core (hidden size 5,280). Top-1 accuracy at steps 1--8: 0.332, 0.320, 0.303, 0.362, 0.383, 0.378, 0.402, 0.438. A probe trained at step~1 and moved to steps 2--8 changes by $+0.017$, $+0.017$, $-0.018$, $-0.060$, $-0.023$, $-0.037$ and $-0.093$; only the changes at steps 5 ($[-0.110, -0.012]$) and~8 ($[-0.142, -0.045]$) exclude zero.

\section{KV-cache branching}\label{app:branching}
Ouro's cache holds one key--value entry per layer and loop, $4\times48=192$ per token. A branch keeps its own cache, positions, attention mask and generated tokens, aligned across all 192 entries at every step. Cached decoding matches full recomputation, bit-exact at prefill and within bf16 rounding when decoding; branches forked, batched, pruned and perturbed stay independent; and a branch denied its own cache diverges (RMS about 3.0). A branch perturbed at loop $u$, layer $\ell$ would normally require recomputing the prompt, because the cache stores keys and values but not the residual stream. I store the residual at that boundary during one shared prefill, add the perturbation, and recompute only layers $\ell+1$ onward on loop $u$ and all later loops. The result matches a full recomputation bit-for-bit in all 192 entries and in the generated continuation, and saves 13--88\% of the per-branch layer computations during prefill, depending on the loop of the fork; I did not measure whole-generation wall-clock time. The branching screen used four tasks, 12 plain samples per task at temperature 0.7 and top-$p$ 0.95, and a 96-token budget.

\section{Steering}\label{app:steering}

\begin{table}[h]
\centering
\caption{Ways of choosing a steering direction, and the outcome on held-out questions. Directions were added through decoder-layer hooks, at layer~36 where recorded, at the last prompt token, with strength $\alpha\le0.02$.}
\label{tab:steering}
\small
\begin{tabular}{@{}lp{0.52\linewidth}@{}}
\toprule
Direction & Outcome \\
\midrule
Probe weight vector & Outputs change; success does not change in the intended direction (6 tasks per mode) \\
Mean difference, successful minus unsuccessful & Same (6 tasks per direction and mode) \\
Probe direction, RMS-calibrated & Same (6 tasks) \\
Gradient of a probe's score & No better than a random direction (18 scored rows) \\
Adapter derived from a classifier & No reliable effect on held-out questions (42 scored rows) \\
Adapter trained under teacher forcing & Raises the logit margin under teacher forcing only; no effect in free generation (8 tasks) \\
Adapter trained with REINFORCE & Raises reward in training; on 12 held-out tasks (4 samples each) 0.479, against 0.458 for a random direction and 0.438 without steering \\
\bottomrule
\end{tabular}
\end{table}

Before any steering I checked that zero-strength hooks reproduce unhooked generation exactly, that effects scale with strength and reach the logits, lasting about 32 tokens, and that there is no numerical instability at the strengths used. For unit-norm directions, $\alpha=0.02$ is an effective perturbation of about 0.0003 in RMS; the RMS-calibrated directions reach 0.016. The REINFORCE adapter also changed the outcome of fewer tasks than the random direction (2 against 5). The probe direction and the mean difference have cosine similarity $+0.101$. For the adapters only proxies exist: the mean of the teacher-forced adapter's up-projection weights has cosine $-0.00055$ with the probe direction and $-0.00429$ with the mean difference, and the two adapters' average outputs on 64 random Gaussian inputs have cosine $+0.951$. These proxies do not measure where the adapters push the model's reasoning states.

\section{Using the signal: protocols}\label{app:use}

\paragraph{Abstention} Preserved predictions (held-out for Horizon, out-of-fold for GSM8K); no new model runs. Areas under the accuracy--coverage curve are compared between rankers on the same answers with 2,000 question-clustered bootstrap draws. Pointwise, accuracy is higher at every recorded coverage below 100\% and equal at 100\%; the interval excludes zero at 7 of the 10.

\begin{table}[h]
\centering
\caption{Gain in area under the accuracy--coverage curve from the probe over surface features alone, between 50 and 100\% coverage and over all coverage (the area under the risk--coverage curve falls by the same amount). The GSM8K row uses the labels of the first GSM8K test (Section~\ref{sec:cut}).}
\label{tab:abstain}
\small
\begin{tabular}{@{}lrll@{}}
\toprule
Comparison & Answers & 50--100\% coverage & All coverage \\
\midrule
Horizon, all questions, probe with four surface features & 614 & $+0.0124$ $[+0.0048, +0.0207]$ & $+0.0287$ $[+0.0143, +0.0462]$ \\
Horizon, all questions, five-feature baseline & 614 & $+0.0119$ $[+0.0044, +0.0196]$ & $+0.0281$ $[+0.0136, +0.0459]$ \\
Horizon, new questions only, five features & 443 & $+0.0093$ $[+0.0006, +0.0184]$ & $+0.0294$ $[+0.0113, +0.0509]$ \\
GSM8K, hidden states only & 680 & $+0.0087$ $[+0.0003, +0.0181]$ & \\
\bottomrule
\end{tabular}
\end{table}

\paragraph{Agreement among samples (added after the results)} Three answer-free runs: the second Horizon set (2,376 answer-free answers from 1,067 questions, 211 incorrect), the GSM8K training split (6,514 answers from 1,879 problems, 295 incorrect) and the first Horizon set (1,339 answers, 101 incorrect). Every answer is scored out of fold, over five folds assigned by a hash of the question; the probe uses only the hidden states at the run's cut.
\begin{itemize}
\item \emph{Baselines.} For single answers: the finished answer's length, mean log-probability, mean token entropy and token-limit flag; the share of parseable answers to the question that give the same answer; whether it is the majority answer; and the question features. For the majority answer, one row per question: mean length, log-probability and token entropy over the four answers, and the share agreeing with the majority answer. The majority rule, the metrics and the bootstrap (1,000 question resamples) reuse code from \url{https://github.com/djaniak/reasoning-geometry} (commit \texttt{26c48be}, MIT license).
\item \emph{Results.} For majority answers the area under the risk--coverage curve changes by $-0.0010$ ($[-0.0042, 0.0014]$) on Horizon and $+0.0008$ ($[0.0000, 0.0017]$) on GSM8K. On the first set, single answers change by $+0.0013$ ($[-0.0013, 0.0054]$); a majority-answer gain there vanishes when the question features are in both arms, so the question features account for it. Among unanimous answers the probe's AUROC is 0.691 on Horizon and 0.594 on GSM8K. Against the window-only baseline plus the window's token entropy, it adds $+0.036$ ($[0.020, 0.053]$) on Horizon, $+0.022$ ($[0.001, 0.042]$) on GSM8K and $+0.005$ ($[-0.019, 0.028]$) on the first set.
\item \emph{Stacking.} The probe's out-of-fold scores enter a second-stage model; the scores of its training rows come from probes that saw the test fold's labels, so the two-stage pipeline is not fully nested. Token entropy was recomputed with one forward pass per answer.
\end{itemize}

\paragraph{Reranking} First pool: the first 170 Horizon questions, split 80/31/59; unparseable answers count as incorrect. The probe reads the hidden states of the prompt plus the complete answer and uses 16 PCA components with $\lambda=4.0$. Exact McNemar test between probe and surface selector, $p=1.00$. Second pool: each question was resampled until it had four parseable answers; 369 questions, 110 held-out groups, 32 informative.

\paragraph{Pruning inside the loop} 48 tasks (24 reasoning, 24 science), at layers 24, 36 and~47 on loops 1--4; two ranking probes, used together, score all candidates pairwise. Of 528 pruning decisions over all 48 tasks, the best-reward branch survived 520 (98.5\%), but 26 of the tasks were on the training side of the probes' split, which also let 8 questions cross. Of the 48 tasks, 13 are on a held-out list declared before the run; one of the two probes had seen 4 of them in training, and on the remaining 9 the best-reward branch survived 96 of 99 decisions. Of 3,406 children, 3.9\% improved on their parent's reward and 3.3\% did worse. In 8 of the 13 tasks whose branches differ, at least 44\% of branches tie for the best reward.

\paragraph{Pruning sampled answers} Six answers per question in two seeded rounds of three (temperature 0.7, top-$p$ 0.95, at most 320 new tokens), all generated to completion; policies were replayed on the saved features after 80 tokens and at the end, with the frozen pairwise ranking probes (layers 24, 36 and~47 on loop~4). Random pruning keeps two samples at random (500 replicates). The preregistered decisions come in order: a correct answer among six on 15--95\% of questions; the 95th percentile of random pruning; then a paired bootstrap against best-of-3 (2,000 rounds). The first run, on proof depths 2--4, failed the first check (97.8\%). The two finished survivors contained a correct answer on 63.9\% of questions, against 75.0\% for the first three samples. The token counts (tournament 868, best-of-3 814, best-of-4 1,076) count generated tokens only. The probe's score at 80 tokens predicted final correctness with AUROC 0.561.

\paragraph{Loops per question} 180 new Horizon questions per checkpoint. Every question was answered twice at each depth from 1 to 4 loops, so every policy selects rows of the same table, and policies are evaluated on the 50 held-out questions. The probe is a ridge regression on the prompt's loop-1 hidden states; the gate policy uses the expected exit step from Ouro's exit gate, averaged over prompt tokens. The random control draws 100 assignments with the same depth histogram, which is identical for the probe and the gate at all 19 budgets; the run's verdict tests the probe. Beyond its one crossing, the gate equals the 95th percentile at three Ouro-RLTT budgets. The matching is on assigned loop depth only. Success at fixed depth 1--4 is 0.38, 0.84, 0.86 and 0.80 for Ouro-RLTT, and 0.18, 0.44, 0.64 and 0.70 for Ouro-2.6B-Thinking.

\paragraph{Light training} The 300-step LoRA used 30.3M trainable parameters (1.12\%); it raised branch diversity from 2.17 to 2.62 and the parse rate in code from 0.72 to 0.94, while pass@$K$ moved from 0.75 to 0.92 in math, 0.83 to 0.75 in logic and 0.92 to 0.67 in reasoning. The binding experiment used LoRA rank 16 on all linear layers, 147 steps (cap 400), and batch size 1 with 8 accumulation steps. The direction was a fixed unit-RMS direction added at layer 24 on loop 1 at the last prompt token with strength 0.02. Evaluation used 90 new questions, four samples each, with paired seeds. Success with and without the direction: trained adapter 0.422 and 0.414, shuffled-sign adapter 0.425 and 0.436, frozen model 0.553 and 0.508. The trained adapter lowered plain accuracy by 9.4 points ($[-0.164, -0.019]$).

\section{Geometry}\label{app:geometry}
The perturbations applied in the branching experiments span a subspace of rank 344 in the 24,576-dimensional feature space, although the participation ratio is only about 5.1. The direction separating correct from incorrect continuations, estimated from 8 correct and 336 incorrect branches, places 0.0183 of its squared norm in this span. A random direction places 0.0140 there in expectation, and the observed value lies above 99.99\% of $10^6$ random directions. Directions fitted to shuffled labels place 0.0230 there on average ($10^4$ draws), and the observed value lies at their 3.4th percentile. The probe and outcome subspaces overlap far beyond chance; neither aligns with the perturbations beyond chance at the three positions where perturbation data exist, and none exist at layers 8 or~16.

\section{Reproducibility}\label{app:repro}

\paragraph{Models} Ouro-2.6B, \texttt{ByteDance/Ouro-2.6B}, revision \texttt{1ed04250da1a9936042725d302e81c8fa2ab5abd}. Ouro-2.6B-Thinking, \texttt{ByteDance/Ouro-2.6B-Thinking}, revision \texttt{f1edd81e7ac41355db670500ceaf204e0f73af68}. Ouro-1.4B, \texttt{ByteDance/Ouro-1.4B}, revision \texttt{574fa66cb8bf5abdc979642d01cf2b79b16bfab1}. Ouro-RLTT: research weights provided by Jonathan Williams, converted locally, with shard hashes recorded; they are not public, so the RLTT results cannot be reproduced end to end, but all derived features, predictions, probes and statistics are preserved with checksums. MiniCPM-2B-sft, \texttt{openbmb/MiniCPM-2B-sft-bf16}, revision \texttt{4ec16344ac13e6ef5010aeecaa533369ac8eb53c}. Huginn, \texttt{tomg-group-umd/huginn-0125}, revision \texttt{bb6621b65e90b6a4b9b29ef88dc83866d450470c}. The text readers: \texttt{microsoft/deberta-v3-large}, revision \texttt{64a8c8eab3e352a784c658aef62be1662607476f}, and \texttt{Qwen/Qwen2.5-1.5B}, revision \texttt{8faed761d45a263340a0528343f099c05c9a4323}.

\paragraph{Environment} \texttt{transformers} 4.54.1 and PyTorch 2.12.0.dev20260407 (CUDA 12.8), on one NVIDIA RTX 5070 Ti Laptop GPU. The second set of Section~\ref{sec:fixed}, its Ouro-2.6B-Thinking run and the GSM8K training-split run were generated on NVIDIA A40 GPUs (Runpod) with \texttt{transformers} 4.54.1 and PyTorch 2.8.0 (CUDA 12.8), from the same Ouro-RLTT weights, checked by SHA-256.

\paragraph{Seeds} Splits 20260711; GSM8K bootstrap 20260710; Section~\ref{sec:loops} 20260720; Horizon generation and analysis 20260724; GSM8K generation at a fixed cut 20260927 (its analysis uses the Horizon seed); GSM8K training split 20261003; loops per question 20260726; pruning of sampled answers and the LoRA binding experiment 20260727.

\paragraph{Records} Code and run records are in \url{https://github.com/VykosMolt/Branching-Looped-Transformer}, under \texttt{results/}, pinned to commit \texttt{0026197381bb92bdceaf515f1a1cc392471f3fad}. The run directories added for this version list every file of their run with SHA-256 checksums, including the feature shards, which are kept offline because of their size. The directories are: \texttt{cross\_loop\_early\_layer\_taps\_20260720}, \texttt{family\_xloop\_v3\_20260726} and \texttt{huginn\_probe\_v3\_20260726} (Section~\ref{sec:loops}); \texttt{paper1\_v2\_overnight\_20260724}, \texttt{horizon\_power\_v3\_20260726}, \texttt{thinking\_preanswer\_v3\_20260726}, \texttt{thinking\_preanswer\_power\_v5\_20260727} and \texttt{fixed\_prefix\_rerun\_20260927} (Section~\ref{sec:correct}); \texttt{visible\_prefix\_hidden\_increment\_20260729T075247Z} and \texttt{fixed\_prefix\_horizon\_20260729T082827Z} (the audit); \texttt{prompt\_only\_horizon\_20260926} (the prompt-only analysis); \texttt{horizon\_cohort2\_20261003} with its \texttt{posthoc\_audit\_20261004}, \texttt{gsm8k\_fixed\_prefix\_20260927}, \texttt{gsm8k\_train\_20261003} and \texttt{opi\_posthoc\_20261003} (the second set, the GSM8K tests, and the checks run after the first set's result); \texttt{selective\_prediction\_v3\_20260726}, \texttt{wellformed\_terminal\_v4\_20260727}, \texttt{depth\_alloc\_v3\_20260726}, \texttt{tournament\_v4\_20260727}, \texttt{tournament\_rerun\_depth4\_20260927} and \texttt{lora\_s3a\_pilot\_20260727} (Section~\ref{sec:use}); \texttt{posthoc\_astra\_20261010} (checks added in response to an outside review); \texttt{horizon\_thinking\_20261010} with \texttt{posthoc\_exitgate\_20261018} and \texttt{posthoc\_run8\_unpaired\_20261018} (the answer-free test on Ouro-2.6B-Thinking); \texttt{posthoc\_review\_20261108} (tie handling for top-1); and \texttt{gsm8k\_rest\_20261010}, a GSM8K plan that was dropped before generation and holds code the Ouro-2.6B-Thinking analysis uses.

\section{Changes from version 3}\label{app:changes}
\begin{itemize}
\item The title is kept with a new subtitle, and its term is now defined narrowly (Section~\ref{sec:intro}). The other coined terms of version~3 (``taps'', ``readout--control boundary'', ``decision-level use'' versus ``generative control'', and others) are replaced by standard ones.
\item Section~\ref{sec:correct} adds answer-free tests run for this version: two Horizon sets, the second fixed before generation with question difficulty in the baseline and each answer labeled by its first final answer; two GSM8K tests; and Ouro-2.6B-Thinking. It adds readers of the visible text and comparisons within questions. The final-answer-line test of earlier versions is kept and is no longer described as strictly pre-answer (Section~\ref{sec:cut}); its combined probe, which left the surface features almost no weight, is replaced by one that compresses the hidden states only.
\item Section~\ref{sec:loops} adds moves of probes to earlier loops. Version~3 concluded that the first loop stores the ranking direction in different coordinates; the moves show that the direction is largely shared across loops, which is most simply read as the first loop holding it weakly. Top-1 accuracy now splits credit among candidates tied at the top score (Appendix~\ref{app:features}); this lowers the probes' values by 0.006--0.008 and puts ranking by length below chance.
\item Section~\ref{sec:use} adds the answer-free probe against a baseline that knows the agreement among finished samples, full-coverage abstention metrics, a sealed rerun of the sampled-answer pruning on 180 harder questions, and a comparison of branch pruning with random pruning.
\item Numbers that differ from version~3: the five-domain ranking probe reaches 0.627 on a split without the 195 colliding IDs (version~3 also reported 0.669, measured on a split those IDs crossed); the branch-pruning survival rate of 97.0\% comes from the 9 held-out tasks, not from all 48, and the share of tasks with many tied branches is 8 of 13. Version~3 described an ablation as showing that the layer-47 channel read by the branch-pruning probe is causally load-bearing; the ablation removed the perturbations that generate many of the branches, so it shows where the best branches come from, not what the probe reads.
\item Left out: the preference probes and the claim that reasoning training installs the preference signal, which had no surviving record and did not reproduce (see \citet{kirin2026rpe}); the domain-transfer study; the construction of the branch-pruning and ranking probes; the details of the generated-answer correctness test on 16 questions, whose result is in Section~\ref{sec:discussion}; the branch-convergence detector; the MATH observation that motivated the project; and speculative applications. All remain in version~3.
\end{itemize}

\endgroup
\end{document}